\documentclass[1p, authoryear]{elsarticle}
\usepackage{macros, rawfonts, amsmath, amssymb, graphicx, algorithm, subcaption, hyperref, tikz, mathrsfs, enumitem, booktabs, hyperref}
\usepackage[noend]{algpseudocode}

\hypersetup{ colorlinks=false, linkcolor=blue, filecolor=magenta, urlcolor=blue}
\usetikzlibrary{positioning,shapes,shapes.geometric,arrows,calc}

\newtheorem{defin}{Definition}
\newtheorem{fact}{Fact}

\newenvironment{defn}{
\begin{defin}\begin{quote}\em}
        {\end{quote}\end{defin}}

\newcommand{\nncite}{\citep}
\newcommand{\namecite}{\cite}

\newcommand{\cpt}{{\sc cpt}}
\newcommand{\dn}{{\sc dn}}
\newcommand{\dns}{{\sc dn}s}
\renewcommand{\dag}{{\sc dag}}

\newcommand{\drl}{{\sc drl}}
\newcommand{\rl}{{\sc rl}}
\newcommand{\sample}{\mathit{Sample}}
\newcommand{\dial}{{\cal D}}
\newcommand{\update}{\mathit{update}}

\newcommand {\lsembrac}         {\mbox{[ \kern -0.5 em [}}
\newcommand {\rsembrac}         {\mbox{] \kern -0.5 em ]}}
\newcommand {\sem} [1]          {{\lsembrac #1 \rsembrac}}

\newcommand{\ra}[1]{\renewcommand{\arraystretch}{#1}}

\newcommand {\csmodels} {\hspace*{0.5em}\mbox{$\mid$ \kern -0.5 em $\approx$}\hspace*{0.5em}}
\newcommand {\csmodel}{\mbox{$\mid$ \kern -0.5 em $\approx$}}

\newcommand {\notcsmodels} {\hspace*{0.5em}\mbox{$\mid$ \kern -0.5 em $\approx$ \kern -1 em $/$}\hspace*{0.5em}}

\newcommand{\csprove} {\mbox{$\mid$ \kern -0.5 em $\sim$}}

\newcommand{\notcsprove} {\mbox{$\mid$ \kern -0.5 em $\sim$ \kern -1 em $/$}}

\newcommand{\csproves} {\hspace*{0.5em}\mbox{$\mid$ \kern -0.5 em     $\sim$}\hspace*{0.5em}} 

\newcommand{\notcsproves} {\hspace*{0.5em}\mbox{$\mid$ \kern -0.5 em
    $\sim$ \kern -1 
    em $/$}\hspace*{0.5em}}

\newcommand{\hidden}[1]{}

\makeatletter
\def\ps@pprintTitle{%
	\let\@oddhead\@empty
	\let\@evenhead\@empty
	\def\@oddfoot{\centerline{\thepage}}%
	\let\@evenfoot\@oddfoot}
\makeatother

\begin{document}
	
\title{Reasoning about Unforeseen Possibilities During Policy Learning}

\author[edi]{Craig Innes}
\ead{craig.innes@ed.ac.uk}

\author[edi]{Alex Lascarides}
\ead{alex@inf.ed.ac.uk}

\author[edi]{Stefano V. Albrecht}
\ead{s.albrecht@ed.ac.uk}

\author[edi]{Subramanian Ramamoorthy}
\ead{sramamoo@inf.ed.ac.uk}

\author[wit]{Benjamin Rosman}
\ead{brosman@csir.co.za}

\address[edi]{School of Informatics, University of Edinburgh}
\address[wit]{Mobile Intelligent Autonomous Systems, CSIR, and School of Computer Science and Applied Mathematics, University of the Witwatersrand}
	
\begin{abstract}
	Methods for learning optimal policies in autonomous agents often assume that the way the domain is {\em conceptualised}---its possible states and actions and their causal structure---is known in advance and does not change during learning. This is an unrealistic assumption in many scenarios, because new evidence can reveal important information about what is possible, possibilities that the agent was not aware existed prior to learning.  We present a model of an agent which both discovers and learns to exploit unforeseen possibilities using two sources of evidence: direct interaction with the world and communication with a domain expert. We use a combination of probabilistic and symbolic reasoning to estimate all components of the decision problem, including its set of random variables and their causal dependencies. Agent simulations show that the agent converges on optimal polices even when it starts out unaware of factors that are critical to behaving optimally.
\end{abstract}

\maketitle
\section{Introduction}
\label{sec:intro}
	
Consider the following decision problem from the domain of crop
farming (inspired by \namecite{kristensen2002use}): Each harvest
season, an agent is responsible for deciding how best to grow barley
on the land it owns. At the start of the season, the agent makes some
decisions about which grain variety to plant and how much fertiliser
to use. Come harvest time, those initial decisions affect the yield
and quality of the crops harvested. We can think of this problem as a
single-stage (or ``one-shot'') decision problem, in which the agent
chooses one action based on a set of observations, then receives a
final reward based on the outcome of its action.

Suppose the agent has experience from several harvests, and believes
it has a good idea of the best seeds and fertiliser to use for a given
climate. One harvest, something totally unexpected happens: Despite
choosing what it thought were the best grains and fertiliser, many of
the crops have come out deformed and of poor quality. A neighbouring
farmer tells the agent the crops have been infected by a fungus that
spreads in high temperatures, and that the best way to protect crops
in future is to apply fungicide. The agent must now revise its model
of the environment and its reward function to account for fungus (a
concept it previously was not aware of), extend its available actions
to include fungicide application (an action it previously did not
realise existed), reason about the probabilistic dependencies
these new concepts have to the ones it was already aware of, and reason
about how they affect rewards.
	
This example illustrates at least three challenges, the combination of
which typically is not handled by current methods for learning optimal
decision policies (we defer a detailed discussion of related work to
Section~\ref{sec:related-work}):
	
\begin{itemize}
\item In addition to starting unaware of the probabilistic dependency
  structure of the decision problem, the agent starts unaware of even
  the true \emph{hypothesis space} of possible problem structures,
  including the sets of possible actions and environment variables,
  and their causal relations.
\item Domain exploration alone might not be enough to discover these
  unknown factors. For instance, it is unlikely the agent would
  discover the concept of fungus or the action of fungicide
  application by just continuing to plant crops. External \emph{expert
    instruction} is necessary to overcome unawareness.
\item An expert might interject with contextually relevant advice
  \emph{during learning}, not just at the beginning of the
  problem. Further, that advice may refer to concepts which are not a
  part of the agent's current model of the domain.
\end{itemize}
	
In the face of such strong unawareness, one might be tempted to
side-step learning an explicit model of the problem, and instead learn
optimal behaviour directly from the data. Deep reinforcement learning
(e.g. \namecite{mnih:etal:2015}) has proved extremely useful for
learning implicit representations of large problems, particularly in
domains where the input sensory streams are complex and
high-dimensional (e.g computer vision). However, in many such
models, the focus of attention for abstraction and representation
learning is on perceptual features rather than causal relations 
(see, e.g., \cite{pearl:17-r475}) and other decision-centric attributes. 
For instance, in works such as by \namecite{chen:etal:2015}, who 
demonstrate an end-to-end solution to the problem of autonomous 
driving, the decisions are not elaborated beyond ``follow the lane'' or 
``change lanes'' although significant perceptual representation learning 
may need to happen in order to work with the predicate ``lane'' within the 
raw sensory streams. For safety critical decisions, or ones involving significant 
investment (e.g. driving a car, advising on a medical procedure, deciding on
crops to grow for the year) it is important \footnote{As evidenced by
significant recent interest from agencies interested in the application
of AI, e.g., the DARPA Explainable AI programme \namecite{darpa_xai:2016}} 
that a system can explain the reasoning behind its decisions 
so the user can trust its judgement.
	
We present a learning agent that, in a complementary approach to
representation learning in batch mode from large corpora, uses
evidence and a reasoning mechanism to incrementally construct an {\em
  interpretable} model of the decision problem, based on which optimal
decision policies are computed. The main contributions of this paper
are:
	
\begin{itemize}
\item An agent which, starting unaware of factors on which an optimal
  policy depends, learns optimal behaviour for single-stage decision
  problems via direct experience and advice from a domain expert. The
  learning uses \emph{decision networks} to represent beliefs about
  the variables and causal structure of the decision problem,
  providing a compact and interpretable model. Crucially, the agent
  can revise \emph{all} components of this model, including the set of
  random variables, their causal dependencies, and the domain of the
  reward function (Section \ref{sec:model}).
\item A communication framework via which an expert can offer both
  solicited and unsolicited advice to the agent \emph{during}
  learning: that is, the expert advice is offered piecemeal, in
  reaction to the agent's latest attempts to solve the task, rather
  than all of it being conveyed prior to learning.  Messages from the expert can
  include entirely new concepts which the agent was previously
  unaware of, and provide important \emph{qualitative} information to
  complement the \emph{quantitative} information conveyed by
  statistical correlations in the domain trials (Section
  \ref{sec:dialogue}).
\item Experiments across a suite of randomly generated decision
  problems, which demonstrate that our agent can learn the optimal
  policy from evidence, even if it were when initially unaware of variables 
  that are critical to its success (Section \ref{sec:experiments}).
\end{itemize}

The kinds of applications we ultimately have in mind for this work
include tasks in which there is a need for flexible and robust
responses to a vast array of contingencies.  In particular, we are
interested in the paradigm of continual (or life-long) learning
\nncite{Silver:2011,thrun:2012}, wherein the agent must continually
and incrementally add to its knowledge, and so revise the hypothesis
space of possible states and actions within which decisions are
made. In this context, there is a need for autonomous model management
\nncite{liebman:2017}, which calls for reasoning about what the
hypothesis space is, in addition to policy learning within those
hypothesis spaces.

\section{The Learning Task}
\label{sec:task}

We consider learning in \emph{single-stage} decision problems. In
these problems, the agent chooses an action based on a set of initial
observations, then immediately receives a final reward based on the
outcome of its action.  Subsequent repetitions of the same decision
problem have mutually independent initial observations, and the
immediate reward depends only on the current action and its
outcome. Solving the problem of unforeseen possibilities for
single-stage scenarios is a necessary first step towards the long-term
goal of extending this work to multi-stage, sequential decision
problems.

To learn an optimal decision policy, the agent must compute
which action will maximise its expected reward, given its observations
of the state in which the action is to be performed. Formally, the
optimal action $\vec{a}^*$ given observations $\vec{e}$ is the action
which maximizes \emph{expected utility}:
	
\begin{equation}
\label{eq:optimal-behaviour}
\vec{a}^*= \arg\max_{\vec{a}\in v({\cal A})} EU(\vec{a}|\vec{e}) 
=  \arg\max_{\vec{a}\in v({\cal A})}\sum_{\vec{s}\in v({\cal C})}
Pr(\vec{s}|\vec{a},\vec{e}){\cal R}(\vec{s})
\end{equation}

Here, $\mathcal{A} = \{A_1, A_2, \dots , A_n \}$ is the set of
\emph{action variables} and $\mathcal{C} = \{C_1, C_2, \dots , C_m \}$
is the set of \emph{chance variables} (or state variables). An
\emph{action} $\vec{a} \in v(\mathcal{A})$ is an assignment to each of
the action variables in $\mathcal{A}$, such that $\vec{a} = (a_1, a_2,
\dots, a_n)$ with $a_1 \in A_1$, $a_2 \in A_2$, etc. (We use $v(Q)$ to
denote the Cartesian product of all sets in $Q$.) Similarly, a
\emph{state} $\vec{s} \in \mathcal{C}$ is an assignment to each chance
variable in $\mathcal{C}$. The reward received in state $\vec{s}$ is
$\mathcal{R}(\vec{s})$.

In our task, the agent faces the extra difficulty of starting unaware
of certain actions and concepts (i.e., chance variables) on which the
true optimal policy depends. Consequently, it begins learning with an
incomplete hypothesis space. It also has an incorrect model of the
domain's causal structure---i.e., there may be missing or incorrect
dependencies. Further, the learning agent will start with an
incomplete or incorrect reward function. The agent's learning task,
then, is to use evidence to converge on an optimal policy, despite
beginning with an initial model defined over an incomplete set of
possible states and actions and incorrect dependencies among them.

Since we are interested in interpretable solutions, our approach is
that the learning agent uses evidence to dynamically construct an {\em
  interpretable model} of the decision problem from which an optimal
policy can be computed.  Formally, we treat this task of learning an
optimal policy as one of learning a {\em decision network} (\dn).
\dn{}s capture preferences with a numeric reward function and beliefs
with a Bayes Net, thereby providing a compact representation of all
the components in equation (\ref{eq:optimal-behaviour}).

\begin{defn}{\textbf{Decision Network}}\\
\label{defn:dn}
A Decision Network (\dn) is a tuple $\langle {\cal C}, {\cal A},
\Pi,\theta,{\cal R}\rangle$, where ${\cal C}$ is a set of chance
variables, ${\cal A}$ is a set of action variables (the agent controls
their values), and ${\cal R}$ is a reward function whose domain is
$\Pi_R\subseteq {\cal C}$ and range is $\mathbb{R}$.\footnote{ Agents
do not have intrinsic preferences over actions, but rather over their
outcomes.}  $\langle \Pi,\theta\rangle$ is a {\em Bayes Net} defined
over ${\cal C}\cup {\cal A}$.  That is, $\Pi$ is a directed acyclic
graph (\dag), defining for each $C\in {\cal C}$ its parents $\Pi_c
\subseteq {\cal C}\cup {\cal A}$, such that $C$ is conditionally
independent of $({\cal C}\cup {\cal A})\setminus \Pi_c$ given
$\Pi_c$. $\theta$ defines for each variable $C\in {\cal C}$ its
conditional probability distribution $\theta_c= Pr(C|\Pi_c)$.
\end{defn}

A policy for a {\sc dn} is a function $\pi$ from the observed portion
$\vec{e}$ of the current state (i.e., $\vec{e}$ is a subvector of
$v({\cal C})$) to an action $v({\cal A})$.  As is usual with \dn{}s,
any variable $X$ whose value depends on the action performed (in other
words, $X$ is a descendant of ${\cal A}$) cannot be observed until
{\em after} performing an action. More formally, ${\cal C}={\cal
  B}\cup {\cal O}$, where the ``before'' variables ${\cal B}$ are
non-descendants of ${\cal A}$, and the ``outcome'' variables ${\cal
  O}$ are descendants of ${\cal A}$ (i.e., where $\Pi^*$ is the closure of
$\Pi$, $X\in {\cal O}$ iff $\exists A\in {\cal A}$ such that $A\in
\Pi^*_X$). So a policy $\pi$ for the \dn{} is a function from $v({\cal
  B}')$ to $v({\cal A})$, where ${\cal B}'\subseteq {\cal B}$ are the
observable variables in ${\cal B}$.

\begin{figure}[t]
	\centering
	\begin{subfigure}[b]{\textwidth}
		\centering
		\includegraphics[scale=0.6]{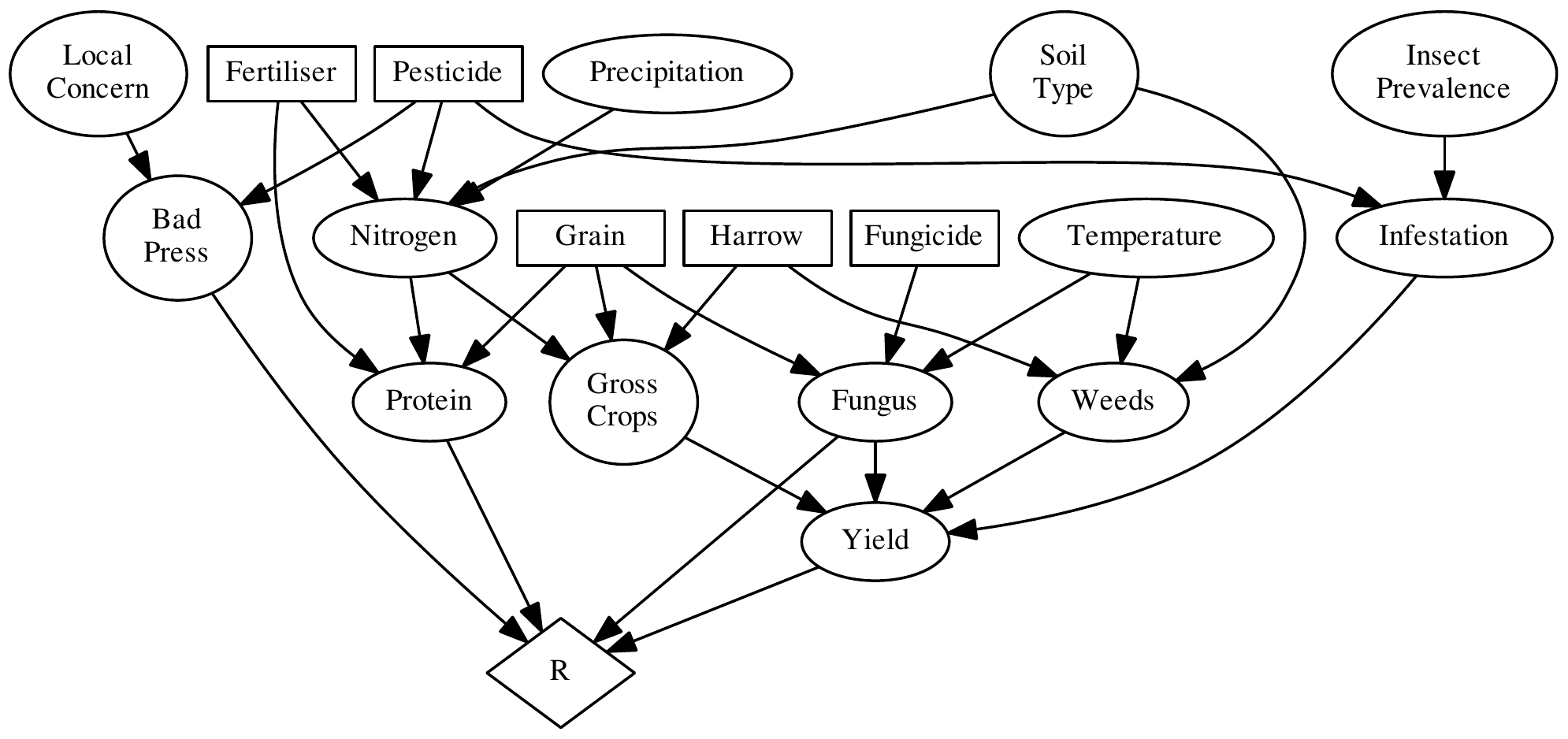}
		\caption{True Decision Problem}
		\label{fig:barley:true}
	\end{subfigure}
	\begin{subfigure}[b]{\textwidth}
		\centering
		\includegraphics[scale=0.6]{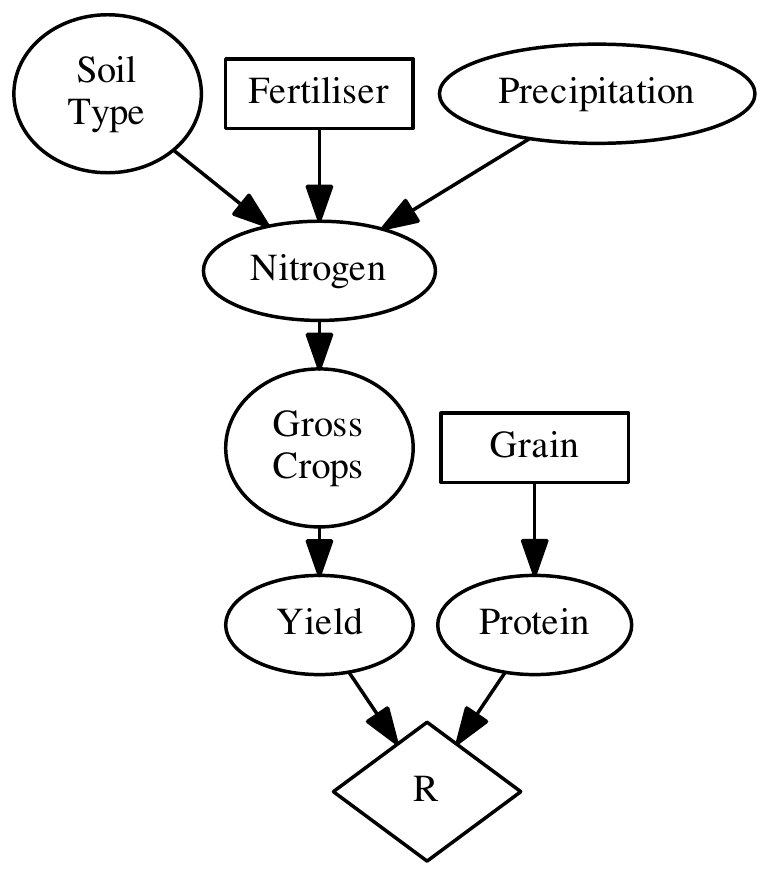}
		\caption{Agent's Initial Model}
		\label{fig:barley:agent}
	\end{subfigure}
	\caption{The graphical component of the \dn{} for the ``Barley'' problem. Action variables are  represented by rectangles, chance variables by ovals, and the reward node by a diamond.} 
	\label{fig:barley}
\end{figure}

Figure \ref{fig:barley:true} shows a \dn{} representation of the
barley example from the introduction. The action variables
(rectangles) are $\mathcal{A} = \{Grain, Harrow, Fungicide,
Fertiliser, Pesticide\}$, the chance variables (ovals) are
$\mathcal{C} = \mathcal{B} \cup \mathcal{O}$ where $\mathcal{B}$
includes $Precipitation$, $Temperature$, $Soil\ Type$ etc.\ and
$\mathcal{O}$ includes $Yield$, $Protein$, $Fungus$, and $Bad\
Press$. The reward domain is $\Pi_\mathcal{R} = \{Yield, Protein,
Fungus, Bad\ Press\}$.

Our \dn{} formulation has some similarities to influence diagrams
\nncite{howard2005influence}. In contrast to DNs, influence diagrams
allow action nodes to have parents via ``information arcs'' from
chance nodes. While this feature is useful to model multi-stage
decision problems, we do not require it here since all action
variables are assigned simultaneously.

Our agent will {\em incrementally} learn {\em all} components of the
\dn{}, including its set of random variables, dependencies and reward
function. We denote the true \dn{} by $dn_+$, and the agent's current
model of the \dn{} at time $t$ as $dn_t$. A similar convention is used
for each component of the \dn{} (so, for instance, $\mathcal{A}_t$ is
the set of action variables the agent is aware of at time $t$). Thus,
we compute an {\em update} function $dn_t = \update(dn_{t-1},e_t)$,
where $e_t$ is the latest body of evidence and the \dn{}s $dn_{t-1}$
and $dn_t$ may differ in any respect. Figure \ref{fig:barley:agent}
gives an example of an agent's possible starting model $dn_0$ for the
barley example.

Notice that $dn_0$ is missing factors that influence the optimal
policy defined by $dn_+$ (e.g. it is unaware of the concept of
fungus). It is also missing dependencies that are a part of $dn_+$
(e.g. it does not think the choice of grain has any influence on the
amount of crops that will grow). One usually assumes that all
variables in a \dn{} are connected to the utility node because a
variable that is not has no effects on optimal behaviour.  So the agent
knows that $dn_+$ is so connected, but $dn_+$'s set of variables,
causal structure and reward function are all hidden and must be
inferred from evidence.

We make four main assumptions to restrict the scope of the learning task:

\begin{enumerate}
\item The agent can {\em observe} the values of all the variables it is
  aware of (so the domain of its policy function at time step $t$ is
  $v({\cal B}_t)$).  However, it cannot observe a chance variable's
  values at times before it was aware of it, even after becoming aware
  of it---i.e. the agent cannot \emph{re-perceive} past domain trials
  upon discovering an unforeseen factor.

\item The agent cannot perform an action it is unaware of. Formally, if
  $X\in {\cal A}_+$ but $X\not\in {\cal A}_t$, then the fully aware
  expert perceives the learning agent's action as entailing $X = 0$
  (plus other values of other variables in ${\cal A}_+$).  This
  differs from the learning agent's own perception of its action: $X =
  0$ is not a part of the agent's representation of what it just did
  because it is not aware of $X$! Once the agent becomes aware of $X$,
  then knowing that inadvertent actions are not possible, it can infer
  all its past actions entail $X = 0$.  This contrasts with {\em
    chance} variables, whose values at times when the agent was
  unaware of them will always be hidden. Clearly, this assumption does
  not hold across all decision problems (e.g. an agent might
  inadvertently lean on a button, despite not knowing the button
  exists). If we wished to lift this assumption, we could simply treat
  action variable unawareness in the same way as we treat chance
  variable unawareness (as described in Section~\ref{sec:model}). 

\item The set of random variables in the agent's initial \dn{} $dn_0$
  is incomplete rather than wrong: that is, ${\cal B}_0\subseteq {\cal
    B}_+$, ${\cal O}_0\subseteq {\cal O}_+$ and ${\cal A}_0\subseteq
  {\cal A}_+$.  Further, the initial domain of ${\cal R}$ is a subset
  of its true domain: i.e., $\Pi^0_R\subseteq \Pi^+_R$.  This
  constraint together with the dialogue strategies in
  Section~\ref{sec:dialogue} simplify reasoning: \dn{} updates may add
  new random variables but never {\em removes} them; and may {\em
    extend} the domain of ${\cal R}$ but never retracts it. However,
  the causal structure $\Pi$ and reward function ${\cal R}$ can be
  revised, not just refined.
		
\item The expert has complete knowledge of the actual decision problem
  $\dn_+$ but lacks complete knowledge of $dn_t$---the learning
  agent's perception of its decision problem at time $t$.  Further, we
  make the expert {\em cooperative}---her advice is always sincere,
  competent and relevant (see Section~\ref{sec:dialogue} for details).
	\end{enumerate}
	
Any competent agent attempting this learning task should obey two key
principles: 
	
\begin{description}
\item [Consistency:] At all times $t$, $dn_t$ should be consistent.
That is, $\Pi_t$ should be a \dag{} defined over the vocabulary
${\cal C}_t\cup {\cal A}_t$, $\theta_t$ should abide by the basic
laws of probability, and ${\cal R}_t$ should be a well-defined
function that captures an asymmetric and transitive preference
relation over ${\cal C}_t$.
\item [Satisfaction:] Evidence is informative about what {\em can}
happen (however rarely), not just informative about likelihood.  At
all times $t$, $dn_t$ should satisfy all the possibilities that are
entailed by the observed evidence so far.
\end{description}

Consistency is clearly desirable, because anything can be inferred
from an inconsistent \dn, making any action optimal.  Satisfaction is
not an issue in traditional approaches to learning optimal policies,
because evidence never reveals a possibility that is not within the
hypothesis space of the learning agent's initial \dn{}.  But in our task,
without Satisfaction, the posterior \dn{} may fail to represent an
unforeseen possibility that is monotonically entailed by observed
evidence.  It would then fail to capture the unforeseen possibility's
effects on expected utilities and optimal policy.  Our experiments in
Section~\ref{sec:experiments} show that when starting out unaware of
factors on which optimal policies depend, a baseline agent that
does not comply with Satisfaction performs worse than an agent that
does.  We will say $dn_t$ is \emph{valid} if it complies with Consistency
and Satisfaction; it is \emph{invalid} otherwise.

A valid \dn{} is necessary, but not sufficient: $dn_t$ should not only
be valid, but in addition its probabilistic component
$\langle\Pi_t,\theta_t\rangle$ should \emph{reflect the relative
  frequencies in the domain trials}.  Section~\ref{sec:model} will
define a \dn{} update procedure with all of these properties. 
	
The set of valid \dn{}s is always unbounded---it is always possible to
add a random variable to a valid \dn{} while preserving its validity.
So in addition to the above two monotonic constraints on \dn{} update,
we adopt two intuitively compelling defeasible principles that make
\dn{} update tractable.  Indeed, the agent needn't enumerate all valid
\dn{}s; instead, \dn{} update uses the defeasible principles to
dynamically construct a single \dn{} from evidence:

\begin{description}
\item [Minimality:] The \dn{} should have minimal complexity. In other
words, its random variables discriminate one possible state from
another, two states have distinct payoffs, and/or two factors are
probabilistically dependent only when evidence justifies this.
\item [Conservativity:] The agent should minimise changes to the
\dn's hypothesis space when observing new evidence.
\end{description}

Minimality is a form of Occam's razor: make the model as simple as possible while accounting for evidence. Conservativity captures the compelling intuition that you preserve as much as possible of what you inferred from past evidence even when you	have to revise the \dn{} to restore consistency with current evidence. Minimality and Conservativity underlie existing symbolic models of commonsense reasoning \nncite{poole:1993,hobbs:etal:1993,alchourron:etal:1985}, the acquisition of causal dependencies \nncite{bramley:etal:2015,buntine:1991} and preference change \nncite{hansson:1995,cadilhac:etal:2015}.

Our final desirable feature is to support {\bf active learning}: to
give the agent some control over the evidence it learns from next,
both from the domain trials (Section~\ref{sec:domain}) and the dialogue content (Section~\ref{sec:dialogue}).

\section{Evidence}
\label{sec:evidence}

We must use evidence to overcome ignorance about how to conceptualise
the domain, not just ignorance about likelihoods and payoffs.  We
regiment this by associating each piece of evidence $e_t$ with a
formula $\delta_t$ that expresses (partial) information about the true
decision problem $dn_+$, where $\delta_t$ follows monotonically from
$e_t$, in the sense that it would be {\em impossible} to observe $e_t$
unless $\delta_t$ is true of the decision problem $dn_+$ that
generated $e_t$.  Thus $\delta_t$ is a {\em partial description} of a
\dn{} that must be satisfied by the actual (complete) decision network
$dn_+$.

We illustrate this with three examples.  First, given the
assumptions we made about the relationship between the agent's initial
\dn{} $dn_0$ and $dn_+$ (i.e., that ${\cal B}_0\subseteq {\cal B}_+$,
${\cal A}_0\subseteq {\cal A}_+$, ${\cal O}_0\subseteq {\cal O}_+$ and
$\Pi^0_R\subseteq \Pi^+_R$), $dn_0$ in Figure~\ref{fig:barley:agent}
yields the partial description $\delta_0$ given in (\ref{eq:delta0}):
\begin{equation}
\label{eq:delta0}
\begin{aligned}
&\{Grain, Fertiliser\} \subseteq {\cal A}_+ \quad\wedge \\
&\{Soil\ Type, Precipitation\} \subseteq \mathcal{B}_+ \quad \wedge \\
&\{Nitrogen, Gross\ Crops, Yield, Protein \} \subseteq {\cal O}_+ \quad \wedge \\
&\{Yield, Protein\} \subseteq \Pi^+_R
\end{aligned}
\end{equation}

The second example concerns domain trials: suppose the agent is in a
state where it experiences the reward $r$ and observes that the $Yield$ variable has value $y$ and the $Protein$ variable has value $p$. Equation (\ref{eq:delta-example}) must be true of $dn_+$, where the domain of quantification is $dn_+$'s atomic states:
\begin{equation}
\label{eq:delta-example}
\exists s(s\rightarrow y \wedge p \wedge {\cal R}_+(s)=r)
\end{equation}
Thirdly, suppose the expert advises the agent to apply pesticide. Then, $Pesticide \in {\cal A}_+$ must be true.

We capture this relationship between a complete \dn{} and formulae
like (\ref{eq:delta0}) and (\ref{eq:delta-example}) by defining a
syntax and semantics of a language for partially describing \dn{}s
(details are in the Appendix).  Each model for interpreting the
formulae $\delta$ in this language corresponds to a unique
complete \dn, and $dn\models \delta$ if and only if $\delta$
(partially) describes $dn$.  Where $\delta_t$ represents the
properties that a \dn{} generating observed evidence $e_t$ must have,
$dn_+\models \delta_t$ (because $dn_+$ generated $e_t$), although
other \dn{}s may satisfy $\delta_t$ too.  This relationship between
$dn_+$, $e_t$ and $\delta_t$ enables the agent to accumulate an
increasingly specific partial description of $dn_+$ as it observes
more and more evidence: where $e_{0:t-1}$ is the sequence of evidence
$e_1,\ldots,e_{t-1}$ and $\delta_{0:t-1}$ its associated partial
description of $dn_+$, observing the latest evidence $e_t$ yields
$e_{0:t}$ with an associated partial description $\delta_{0:t}$ that
is the {\em conjunction} of $\delta_{0:t-1}$ and $\delta_t$. The agent will thus estimate a {\bf valid} \dn{} from evidence (as	defined in Section~\ref{sec:task}) under the following conditions:

\begin{enumerate}
	\item $\delta_{0:t}$ captures the necessary properties of a \dn{} that generates $e_{0:t}$
	\item The agent's model obeys this partial description. That is, $dn_t \models \delta_{0:t}$
\end{enumerate}

This section describes how we achieve the first condition; Section~\ref{sec:model} defines how we
achieve the second, with Section~\ref{sec:pi} also ensuring that
the probabilistic component $\langle \Pi_t,\theta_t\rangle$ reflects
the relative frequencies in the domain trials.

\subsection{Domain Evidence: $\sample_{0:t}$}
\label{sec:domain}

Domain evidence consists of a set of {\em domain trials}. In a domain trial $\tau_i$, the agent observes $\vec{b}_i\in v({\cal B}_i)$, performs an action $\vec{a}_i\in v({\cal A}_i)$, and observes its outcome $\vec{o}_i\in v({\cal O}_i)$ and reward $r_i$. From now on, for notational convenience, we may omit the vector	notation, or freely interchange vectors with conjunctions of their values. Each domain trial therefore consists of a tuple:

\begin{equation}
\label{eq:sample}
\sample_{0:t} = [\langle b_i,a_i,o_i,r_i\rangle: 0 < i\leq t]
\end{equation}

The domain trial $\tau_i=\langle b_i,a_i,o_i,r_i\rangle$ entails the
{\em partial description} (\ref{eq:valid-domain}) of $dn_+$: in words,
there is an atomic state $s\in v({\cal C}_+\cup{\cal A}_+)$ in $dn_+$
that entails the observed values $b_i$, $a_i$ and $o_i$ and which has
the payoff $r_i$.
\begin{equation}
\label{eq:valid-domain}
\exists s((s\rightarrow
(b_i \wedge a_i \wedge o_i)) \wedge {\cal R}_+(s)=r_i)
\end{equation}
Formula (\ref{eq:valid-domain}) follows monotonically from $\tau_i$
because the agent's perception of a domain trial is incomplete but
never wrong: the agent's random variables have observable values, and
they are always a subset of those in $dn_+$ thanks to the agent's
starting point $dn_0$ and the dialogue strategies (see
Section~\ref{sec:dialogue}).  Thus, even if the agent subsequently
discovers a new random variable $X$, (\ref{eq:valid-domain}) still
follows from $\tau_i$, regardless of $X$'s value at time $i$ (which
remains hidden to the agent).  But on discovering $X$, tuples in
(\ref{eq:sample}) get {\em extended}---the agent will observe $X$'s
value in subsequent domain trials.  Thus in contrast to standard
domain evidence, the size of the tuples in (\ref{eq:sample}) is {\em
dynamic}.  We discuss in Section~\ref{sec:model} how the agent copes
with these dynamics.  The expert also keeps a record $\sample^+_{0:t}$
of the domain trials; these influence her dialogue moves (see
Section~\ref{sec:dialogue}).  Her representation of each trial is
at least as specific as the agent's because she is aware of all
the variables---so the size of the tuples in $\sample^+_{0:t}$ is
static.

Domain trials can reveal to the agent that its conceptualisation of
the domain is deficient.  If there are two trials in $\sample_{0:t}$
with the same observed value for $\Pi^t_R$ (i.e., the current
estimated domain of the reward function) but the rewards are
different, then this entails that $\Pi^t_R$ is {\em invalid}.  If
$\sample_{0:t}$ contains two domain trials with the same observed
values for {\em every} chance variable ${\cal C}_t$ of which the agent
is currently aware, but the rewards are different, then the vocabulary
${\cal C}_t$ is invalid.  Section~\ref{sec:model} will define how the
agent detects and learns from these circumstances.

The agent's strategy for choosing an action mixes exploitation and
exploration in an $\epsilon$-greedy approach: In a proportion $(1-\epsilon)$ of the trials, the agent chooses what it currently thinks is an optimal action. In the remainder, the agent chooses an action at random.

\subsection{Dialogue Evidence: $\dial_{0:t}$}
\label{sec:dialogue}

The interaction between the agent and the expert consists of the agent
asking questions that the expert then answers, and unsolicited advice
from the expert. All the dialogue moves are about $dn_+$, and the
signals are in the formal language for partially describing \dn{}s
(see the Appendix), but with the addition of a sense ambiguous term,
which we motivate and describe shortly.  The agent's and expert's
lexica are different, however: the agent's vocabulary lacks the random
variables in $dn_+$ that it is currently unaware of, and so the
expert's utterances may feature {\em neologisms}.  As we said earlier,
we bypass learning how to ground neologisms (but see
\namecite{larsson:2013,forbes:etal:2015,yu:etal:2016}) by assuming
that once the agent has heard a neologism, it can observe its
denotation in all subsequent domain trials.

The {\em sense ambiguous} term is $w^b_t$: its intended denotation is
what the expert observes about the state at time $t$ {\em before} the agent
acts---i.e., $\sem{w^b_t}\in v({\cal B}_+)$.  But this denotation is
hidden to the agent, whose {\em default} interpretation is
$\sem{w^b_t}$ projected onto ${\cal B}_t$---i.e., it is restricted by
the agent's conceptualisation of the domain at time $t$.  The expert
uses $w^b_t$ to advise the agent of a better action than the one it
performed at $t$. We will see in Section~\ref{sec:dialogue:expert}
that using $w^b_t$ in the expert's signal minimises its neologisms, which makes
learning more tractable.  But hidden messages may create
misunderstandings; Section~\ref{sec:dialogue:agent} describes how
the agent detects and learns from them.

The agent and expert both keep a {\em dialogue history}: $\dial_{0:t}$
for the agent and $\dial^+_{0:t}$ for the expert.  Each utterance in a
dialogue history is a tuple $\langle \omega_i,\sigma_i,\mu_i\rangle$, where
$\omega_i$ is the speaker (i.e., either the expert or the
learning agent), $\sigma_i$ the signal, and $\mu_i$ its (default)
interpretation.  Since $w^b_i$ may be misinterpreted, $\mu_i$ and
$\mu^+_i$ may differ and so $\mu_i$ is not equivalent to
$\delta_i$---$\delta_i$ is the information about $dn_+$ that follows
from the signal $\sigma_i$ {\em whatever} the denotation of $w^b_i$
might be.  We will specify $\delta_i$ for each signal $\sigma_i$ in
Section~\ref{sec:dialogue:expert}.

\subsubsection{The expert's dialogue strategy}
\label{sec:dialogue:expert}

As noted earlier, the expert's dialogue strategy is {\em Cooperative}: the message $\mu^+$ that she intends to convey with her signal $\sigma$ is satisfied by the actual decision problem $dn_+$, so that $dn_+\models \mu^+$. This makes her sincere, competent and relevant. In many realistic scenarios, this assumption may be untrue (a human teacher, for example, might occasionally make mistakes). We intend to explore relaxations of this assumption in future work.

Further, the expert's dialogue strategy limits the amount of information she is allowed to send in each signal. There are two motivations for limiting the amount of information. The first stems from the definition itself of our learning task; and the second is practical. First, recall from Section~\ref{sec:intro} that we allow the expert advice to occur piecemeal, with signals being interleaved among the learner's domain trials. This is because a major aim of our learning task is to
reflect the kind of teacher-apprentice learning seen between humans, where the teacher only occasionally interjects to say things that relate to the learner's latest attempts to solve the task. There are many tasks where an expert may be incapable of exhaustively expressing everything they know about the problem domain, but rather can only express relevant information in reaction to specific contingencies that they experience.

The second, more practical motivation is to make learning tractable.
The number of possible causal structures $\Pi$ is hyperexponential in
the number of random variables \nncite{buntine:1991}.  Prior work
utilises defeasible principles such as Conservativity
\nncite{bramley:etal:2015} and Minimality \nncite{buntine:1991} to
make inferring $\Pi$ tractable.  However, if the expert's signal $\sigma_t$
features a set ${\cal N}$ of variables that the agent was unaware of,
then the agent must {\em add} each variable in ${\cal N}$ to the
causal structure $\Pi_t$, and moreover by Consistency and
Satisfaction, each of these must be connected to $\Pi_t$'s utility
node.  But $\Pi_{t-1}$ did not feature these variables at all, and so
the number of maximally conservative and minimal updates that satisfy
this connectedness is hyperexponential in the size of ${\cal N}$.
Thus, an expert utterance with many neologisms undermines the
efficiency and incrementality of learning.

We avoid this complexity by restricting all expert signals to containing {\em at
  most one neologism}.  Specifically, the expert must have conclusive
evidence that the agent is aware of {\em all but one} of the random
variables that feature in her signal $\sigma$.  The expert knows the
agent is aware of a variable $X$ if: (a) $X$ has already been
mentioned in the dialogue, either by her or by the agent; or (b) $X$
is an action variable and the agent has performed its positive value
$x$ (recall that inadvertent action is not possible):
\begin{equation}
	\label{eq:aware-vars}
	\begin{array}{rcl}
		{\cal C}_e & = & \{ X \in {\cal C}_+ \ |\ \exists \sigma_i \in \dial^+_{0:t} : X \in \sigma_i \} \\[5pt]
		{\cal A}_e & = & \{ X \in {\cal A}_+ \ |\ \exists \sigma_i \in \dial^+_{0:t} : X \in \sigma_i \text{ or } \exists \tau_i\in \sample^+_{0:t} : x \in \tau_i \}
	\end{array}
\end{equation}

Here, ${\cal C}_e$ and ${\cal A}_e$ respectively are the set of chance and action
variables that the expert knows the agent is aware of. The following
principle, which we refer to as 1N, applies to all the expert's
signals:

\begin{description}
	\item [At Most One Neologism (1N):] Each expert signal $\sigma$ features at most one variable from $( {\cal A}_+\cup {\cal C}_+ ) \setminus ( {\cal A}_e\cup {\cal C}_e )$ (i.e., at most one neologism). Furthermore, if $\sigma$ features such a variable $X$, then $\sigma$ declares its {\em type}: i.e., $\sigma$ includes the conjunct $X\in {\cal B}_+$, $X\in {\cal O}_+$ or $X\in {\cal A}_+$, as appropriate.
\end{description}

The expert uses an ambiguous term $w^b_t$ to comply with Cooperativity and 1N in contexts where, without $w^b_t$, she would be unable express any advice. For instance, assume that the expert's knowledge of the agent's ``before'' vocabulary is given by $\mathcal{B}_e$. The expert would typically be unable to express that, given $b_t\in v({\cal B}_+)$, the agent should have performed an alternative action $a'$ to the action $a_t$ that the agent actually performed---by 1N, she cannot use the vector $b^+_t \in v({\cal B}_+)$ in her signal if $|{\cal B}_+|-|{\cal B}_e|>1$. If instead she were to use the vector $b^e_t$, where $b^e_t=b^+_t \upharpoonright {\cal B}_e$ (i.e., $b^+_t$ projected onto ${\cal B}_e$), then the resulting statement (\ref{eq:false}) may be {\em false} (and so violate Cooperativity), because these expected utilities marginalise over all possible values of ${\cal B}_+\setminus {\cal B}_e$, rather than using their actual values:
\begin{equation}
	\label{eq:false}
	EU(a'|b^e_t) > EU(a_t|b^e_t)
\end{equation}
Alternatively, by replacing $b^e_t$ in the signal (\ref{eq:false})	with the ambiguous term $w^b_t$, her intended message becomes hidden	but it abides by both Cooperativity and 1N.

The expert's dialogue policy is to answer all the agent's queries as and when they arise, and to occasionally offer unsolicited advice about a better action.  She does the latter when two conditions hold:

\begin{enumerate}[label=(\roman*)]
\item The agent has been behaving sufficiently poorly to justify the
need for advice
\item The current context is one where she can
express a better option while abiding by Cooperativity and 1N.
\end{enumerate}

Condition (i) is defined via two parameters $\gamma$ and $\beta$: in words, the last piece of advice was offered greater than $\gamma$ time steps ago, and from then until now, the fraction of suboptimal actions taken by the agent is greater than $\beta$.  This is formalised in equation (\ref{eq:condition-for-advice-proportion}), where $t'$ is the time of the last advice and $a^{+,*}_i$ is the optimal action given $dn_+$ and $b^+_i$. In the experiments in Section~\ref{sec:experiments}, we vary $\gamma$ and $\beta$ to test how changing the expert's penchant for offering unsolicited advice affects the learning agent's convergence on optimal policies. Condition (ii) for giving advice is satisfied when the observed reward $r_t$ is no higher than the expected payoff from the agent's action $a^+_t$ (equation (\ref{eq:eu-r})), and there is an alternative action $a'$ with a higher expected payoff, which can be expressed while complying with Cooperativity and 1N (equation (\ref{eq:condition-for-advice-eu})).

\begin{equation}
\label{eq:condition-for-advice-proportion}
|t-t'| > \gamma \wedge \frac{|\{a^+_i: t'\leq i\leq t \mbox{ and } EU(a^+_i|b^+_i) <
EU(a^{+,*}_i|b^+_i)\}|}{
|t-t'|} > \beta
\end{equation}

\begin{equation}
\label{eq:eu-r}
EU(a^+_t|b^+_t) \geq r_t
\end{equation}

\begin{equation}
\label{eq:condition-for-advice-eu}
\exists A\exists a'(a'\in v({\cal A}_e \cup
\{A\}) \wedge
EU(a'|b^+_t) \geq EU(a^+_t\upharpoonright ({\cal A}_e\cup \{A\})|b^+_t))
\end{equation}

When the context satisfies these conditions, then there are witness
constants $A$ and $a'$ that satisfy
(\ref{eq:condition-for-advice-eu}).  These constants are used to
articulate the advice: the expert utters (\ref{eq:optimal-action}),
where $a_t$ is the expression formed by projecting the vector
$a^+_t\in v({\cal A}_+)$ onto ${\cal A}_e\cup \{A\}$.
\begin{equation}
	\label{eq:optimal-action}
	EU(a'|w^b_t) > EU(a_t|w^b_t)
\end{equation}
In our Barley example, the message (\ref{eq:optimal-action}) might be
paraphrased as: in the current circumstances, it would have been
better to apply pesticide and not use fertiliser than to not apply
pesticide and use fertiliser. 

The agent's default interpretation of (\ref{eq:optimal-action}) is
(\ref{eq:default-optimal-action}), where $b_t\in
v({\cal B}_t)$:
\begin{equation}
\label{eq:default-optimal-action}
\sum_{s\in {\cal C}_t\times {\cal A}_t} Pr(s|a',b_t){\cal R}_+(s) \ \geq
\sum_{s\in {\cal C}_t\times {\cal A}_t} Pr(s|a_t,b_t){\cal
R}_+(s)
\end{equation}
So $\langle
\mathit{expert},(\ref{eq:optimal-action}),(\ref{eq:default-optimal-action})\rangle$
is added to $\dial_{0:t}$.  While the intended message of
(\ref{eq:optimal-action}) is true (because
(\ref{eq:condition-for-advice-eu}) is true),
(\ref{eq:default-optimal-action}) may be false, 
and so no monotonic
entailments about probabilities can be drawn from it.  For example, in
our Barley example, suppose that the agent's current model of the
domain is Figure~\ref{fig:barley}b, and the agent has observed soil
type $n$ and 
precipitation $c$.  Then the agent's defeasible interpretation of the
expert's advice is that it would have been better to apply pesticide ($p$)
and not use fertiliser ($\neg f$) (than to not apply pesticide and use
fertiliser) in any state where $n\wedge c$ is true (note
that relative to the \dn{} shown in Figure~\ref{fig:barley}b, the
expert mentioning applying pesticide leads it to discover this
entirely new action): 
\begin{equation}
\label{eq:barley-misunderstanding}
\sum_{s\in {\cal C}_t\times {\cal A}_t} Pr(s|p,\neg f,n,c){\cal R}_+(s) \ \geq
\sum_{s\in {\cal C}_t\times {\cal A}_t} Pr(s|\neg p,f,n,c){\cal
R}_+(s)
\end{equation}
But this (defeasible) interpretation
could be false: the expert's (true) intended message may have been
that $p \wedge \neg f$ is better in a much more 
specific situation: one where not only is $n\wedge c$ true, but also
the local concern is low and the insect prevalence is high (in other
words, the probabilities in (\ref{eq:barley-misunderstanding}) should
have been conditioned on $\neg l$ and $i$ as well).  However,
(\ref{eq:optimal-action}) and the mutually known dialogue policy
monotonically entails (\ref{eq:optimal-action-info}), {\em whatever}
the true referent for $w^b_t$ might be.
\begin{equation}
\label{eq:optimal-action-info}
\exists s((s\rightarrow
(a'\wedge b_t)) \wedge {\cal R}_+(s)>r_t)
\end{equation}
So (\ref{eq:optimal-action-info}) is added to $\delta_{0:t}$, and for
$dn_t$ to be valid it must satisfy it; similarly for $A\in {\cal
A}_+$.\footnote{Our learning algorithms in Section~\ref{sec:model}
do not exploit the (defeasible) information about likelihood that is
expressed in (\ref{eq:default-optimal-action}); that is a matter for
future work.}  In our example, the agent adds (\ref{eq:barley-monotonic}) to
$\delta_{0:t}$: 
\begin{equation}
\label{eq:barley-monotonic}
\exists s((s\rightarrow
(p \wedge \neg f\wedge n \wedge c)) \wedge {\cal R}_+(s)>r_t)
\end{equation}
Thus the expert's unsolicited advice can result in the agent
discovering an unforeseen action term $A$ (in this case, adding
pesticide) and/or prompt a revision to 
the reward function ${\cal R}_t$, which in turn may reveal to the
agent that its conceptualisation of the domain is deficient (just as
(\ref{eq:valid-domain}) may do).  

\subsubsection{The agent's dialogue strategy}
\label{sec:dialogue:agent}

\begin{figure}[t]
\centering
\tikzstyle{process} = [rectangle, aspect=1, text centered, draw=black, text width=1cm]
\tikzstyle{decision} = [diamond, aspect=1, text width=0.8cm, text centered, draw=black]
\tikzstyle{arrow} = [thick,->,>=stealth]
\tikzstyle{evidence} = [ellipse,text centered, text width=1.2cm, draw=black]
\tikzstyle{edge} = [thick,-,>=stealth]

\begin{tikzpicture}
\node(dialogue)[evidence]{\scriptsize{dialogue
evidence}};
\node(domain)[evidence, above = 3cm of dialogue]{\scriptsize{domain evidence}};
\node(misunderstanding)[decision, above right=0.3cm of
dialogue]{\hspace*{-0.5cm}\scriptsize{\begin{tabular}{c}misunder-\\standing?\end{tabular}}};
\node(ask-b)[process,below=0.7cm of misunderstanding]{\scriptsize{ask
(\ref{eq:missing-B2})}};
\node(dummy3)[left =0.45cm of ask-b]{};
\node(dummy5)[on grid, below right = 0.4cm and 0.2cm of dialogue]{};
\node(vocab)[process, above right = 0.5cm of
misunderstanding]{\scriptsize{estimate\\ ${\cal 
C}_t$, ${\cal A}_t$, $\Pi^t_R$, ${\cal R}_t$}};
\node(unforeseen)[decision, right=0.4cm of
vocab]{\hspace*{-0.5cm}\scriptsize{\begin{tabular}{c}unforeseen
\\rewards?\end{tabular}}};
\node(pi)[process,right=0.5cm of unforeseen]{\scriptsize{estimate\\  $\Pi_t$}};
\node(ask-reward)[process,below = 0.5cm of unforeseen]{\scriptsize{ask
(\ref{eq:question-unforeseen-reward})}};
\node(dummy7)[below =2.0cm of ask-reward]{};
\node(dummy8)[below =1.14cm of dialogue]{};
\node(valid-pi)[decision,right=0.5cm of pi]{\hspace*{-0.35cm}\scriptsize{\begin{tabular}{c}unknown\\
effects?\end{tabular}}};
\node(ask-effects)[process,below=0.5cm of valid-pi]{\scriptsize{ask
(\ref{eq:which-effect})}}; 
\node(estimate-theta)[process,right=0.5cm of
valid-pi]{\scriptsize{estimate\\ $\theta_t$}};
\node(dummy1)[below =2.25cm of ask-effects]{};
\node(dummy2)[on grid, below left=2.1cm and 0.2cm of dialogue]{};
\node(dummy6)[on grid, below left = 0.4cm and 0.2cm of dialogue]{};
\draw[arrow](domain) -- (vocab);
\draw[arrow](dialogue) -- (misunderstanding);
\draw[arrow](misunderstanding) -- node[sloped, above]{\scriptsize{no}}(vocab);
\draw[arrow](vocab) -- (unforeseen);
\draw[arrow](unforeseen) -- node[anchor=south]{\scriptsize{no}} (pi);
\draw[arrow](unforeseen) -- node[anchor=east]{\scriptsize{yes}} (ask-reward);
\draw[arrow](misunderstanding) -- node[anchor=west]{\scriptsize{yes}} (ask-b);
\draw[arrow](pi) -- (valid-pi);
\draw[arrow](valid-pi) -- node[anchor=south]{\scriptsize{no}}
(estimate-theta);
\draw[arrow](valid-pi) --
node[anchor=east]{\scriptsize{yes}}(ask-effects);
\draw[edge,shorten >=-1.2mm](ask-effects) -- (dummy1);
\draw[edge,shorten <=-1.2mm, shorten >=-1.4mm](dummy1) -- (dummy2);
\draw[arrow,shorten <=-1.2mm, shorten >=1mm](dummy2) -- (dummy6);
\draw[edge,shorten >=-1.0mm](ask-b) -- (dummy3);
\draw[arrow,shorten <=-1.50mm, shorten >=1.0mm](dummy3) -- (dummy5);
\draw[edge,shorten >=-1.0mm](ask-reward) -- (dummy7);
\draw[edge,shorten <=-1.5mm, shorten >=-1.4mm](dummy7) -- (dummy8);
\draw[arrow,shorten <=-1.1mm](dummy8) -- (dialogue);
\end{tikzpicture}

\caption{Information flow when updating a \dn{} with the latest
evidence.  Diamonds are tests and rectangles are processes.}
\label{fig:architecture}
\end{figure}
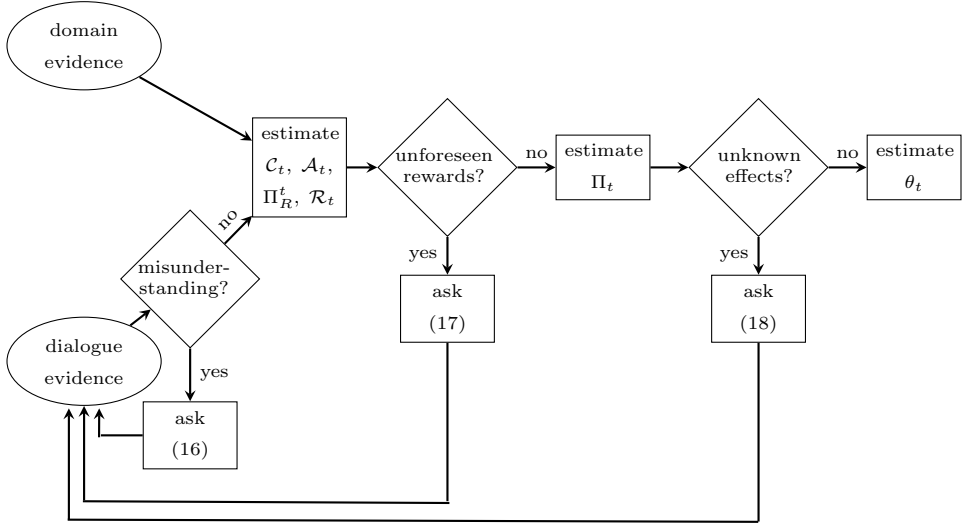

The agent aims to minimise the expert's effort during the learning
process, and so asks a question only when \dn{} update fails to
discriminate among a large number of \dn{}s.  In this paper, we
identify three such contexts (as shown in
Figure~\ref{fig:architecture}):
(i) Misunderstandings; (ii) Unforeseen Rewards; and (iii) Unknown Effects.
We now describe each of these in turn.

\paragraph{Misunderstandings}

The agent checks whether the (default) interpretation of the current signal is consistent with those of prior signals; when they are inconsistent, the agent knows there has been a misunderstanding (because of Cooperativity). For instance, suppose ${\cal B}_t=\emptyset$---so the agent assumes $\sem{w^b_i}=\top$ for all $i\preceq t$---and the expert advised $EU(a|w^b_{t-n})> EU(a'|w^b_{t-n})$ and $EU(a'|w^b_t)>EU(a|w^b_t)$. The agent's default interpretations of these signals fails the consistency test.  Thus the agent infers that it is unaware of a ${\cal B}$ variable, but does not know its name.  If the agent were to {\em guess} what variable to add, then learning would need to support {\em retracting} it on the basis of subsequent evidence, or reason about when it is identical to some subsequent factor the agent becomes aware of.  This is a major potential complexity in reasoning and learning.  We avoid it by defining a dialogue strategy that ensures that the agent's vocabulary is always a subset of the true vocabulary.  Here, that means the agent asks the expert for the name of a ${\cal B}$ variable.  In words, question (\ref{eq:missing-B2}) expresses: what is different about the state before I acted just now and the state before I acted $n$ time steps ago?\footnote{Strictly speaking, answers to this question are partial descriptions of the domain trials $\sample^+_{0:t}$ as well as $dn_+$.  The formal details of this are straightforward but tedious, so we gloss over it here.}

\begin{equation}
\label{eq:missing-B2}
?\lambda V(V\in {\cal B}_+ \wedge \mbox{{\em value}}(V,t-n)\neq
\mbox{{\em value}}(V,t))
\end{equation}

This signal uses a standard representation for questions \nncite{groenendijk:stokhof:1982}: $?$ is an operator that maps a function to a set of propositions (the true answers), and the function's arguments correspond to the {\em wh}-elements of the question (which, what etc).  Its semantics, given in the Appendix, follows \nncite{asher:lascarides:2003}: a proposition is a true answer if and only if it substitutes specific individual(s) for the $\lambda$-term(s) to create a true proposition---so ``something'' is not an answer but ``nothing'' is an answer, corresponding to the empty set.  An answer can thus be {\em non-exhaustive}---it needn't include {\em all} the referents that satisfy the body of the $\lambda$-expression.

The agent's policy for when to ask (\ref{eq:missing-B2}) guarantees that it has a true positive answer.  By the 1N rule, the expert's answer includes one variable, chosen at random if there is a choice.  E.g., she answers with the signal $X\in {\cal B}_+$; so $X\in {\cal B}_+$ is added to $\delta_{0:t}$.  In our barley example, the answer might be, for instance, ``the temperature was high'', where the concept ``temperature'' is a neologism to the learning agent.

The agent now knows that its default interpretations of prior signals
that feature $w^b_i$ ($i\prec t$) are incorrect---on learning $X\in
{\cal B}_+$, it now knows that these terms do not denote $\top$.  But
the agent cannot observe the correct interpretation.  Testing whether
subsequent messages are consistent with prior messages would thus
involve reasoning about past hidden values of $X$, which would be
complex and fallible.  For the sake of simplicity, we avoid it: if the
agent discovers a new ${\cal B}$-variable at time $t$, then she ceases
to test whether subsequent messages $\mu_{t+m}$ are consistent with
past ones $\mu_{t-m'}$ ($m,m'\geq 0$).  In effect, the agent ``forgets''
past signals and their (default) interpretations, but does not forget
their monotonic consequences, which are retained in $\delta_{0:t}$.

\paragraph{Unforeseen rewards}  The current reward function's domain,
$\Pi^t_R$, 
may be too small to be valid (see Sections~\ref{sec:task}
and~\ref{sec:domain}).  If the agent were to guess what variable to
add to $\Pi^t_R$, then learning would need to support {\em retracting} it on the
basis of subsequent evidence---this is again a major potential
complexity in reasoning and learning. Instead, when the agent infers
$\Pi^t_R$ is too small (we define how the agent infers this in
Section~\ref{sec:model}), the agent seeks monotonic evidence for
fixing it by asking the expert (\ref{eq:question-unforeseen-reward}):
	
\begin{equation}
\label{eq:question-unforeseen-reward}
\begin{array}{r l}
?\lambda V(V\in \Pi^+_{R} \displaystyle\bigwedge_{X\in \Pi^t_R}
V\neq X)
\end{array}
\end{equation}

For instance, in the barley domain, this question could express
something like ``Other than having a high yield and high protein
crops, what else do I care about?'' to which the answer might be ``You
care about bad publicity''.  This is a true (non-exhaustive) answer,
even though the
agent (also) cares about avoiding outbreaks of fungus.

Non-exhaustive answers enable the expert to answer the query while
abiding by Cooperativity and the 1N rule.  But they also generate a
{\em choice} on which answer to give.  The expert's choice is driven
by a desire to be informative: she includes in her answer all
variables in $\Pi^+_R$ that she knows the agent is aware of (i.e., the
variables $Y\in {\cal C}_e\cap \Pi^+_R$) and one potential neologism
(i.e., $X\in ({\cal C}_+\setminus {\cal C}_e)\cap\Pi^+_R$) if it
exists, with priority given to a variable whose value is different in
the latest trial $\tau_t$ from a prior trial $\tau_{t'}$ that had the
same values as $\tau_t$ on all the variables mentioned in the query
(\ref{eq:question-unforeseen-reward}) but a different reward.  If her
answer includes a potential neologism, then by 1N she declares its
type.

For example, let $\Pi^{t-1}_R=\{O_1,O_2\}$ and
$\Pi^+_R=\{O_1,O_2,O_3,O_4,O_5\}$.  Suppose that $O_3\in {\cal C}_e$
but $O_4,O_5\not\in {\cal C}_e$.  Now suppose that the latest trial
$\tau_t$ entails $o_1,\neg o_2$ and reward of 1, but a prior trial
entails $o_1,\neg o_2$ and a reward of 0.5.  Then {\em even if} $O_3$
has different values in these trials, the agent does {\em not} infer
(defeasibly) that $O_3\in \Pi^+_R$.  Rather, she asks
(\ref{eq:question-unforeseen-reward}): i.e., $?\lambda V(V\in \Pi^+_R
\wedge V\neq O_1\wedge V\neq O_2)$.  The expert's answer includes
$O_3$ because $O_3\in {\cal C}_e \cap\Pi^+_R$.  In addition, she can
mention either $O_4$ or $O_5$, but not both (because of 1N).  Suppose
that $\tau_t$ entails $o_4$ and a prior trial with a different reward
entails $o_1\wedge \neg o_2 \wedge \neg o_4$.  However, $O_5$ lacks
this property.  Then her answer is: $O_3\in \Pi^+_R \wedge O_4\in
\Pi^+_R \wedge O_4\in {\cal O}_+$.

The ambiguous term $w^b_t$ is not a part of the expert's answer, and so
the message is observable---indeed, it is the same as the signal.  This
becomes a conjunct in $\delta_{0:t}$, and so in our example, for
$dn_t$ to be valid it must satisfy $O_4\in \Pi^t_R$ and $O_4\in {\cal
O}_t$ (and so $\Pi^t_R\supset \Pi^{t-1}_R$ and if $O_4$ truly was a
neologism to the agent then ${\cal O}_t\supset
{\cal O}_{t-1}$ as well).  Section~\ref{sec:model} defines how to estimate a
valid $dn_t$ from this evidence.

\paragraph{Unknown Effects}
There are contexts where the search space of possible causal
structures remains very large in spite of the defeasible principles
for restricting it, in which case the agent asks a question whose
answer will help to restrict the search space:
\begin{equation}
\label{eq:which-effect}
?\lambda V(X\in \Pi_V)
\end{equation}
In words, what does $X$ affect?  For instance, in the barley domain,
it might express: What does the temperature affect? (to which an
answer might be {\em the risk of weeds}).  In Section~\ref{sec:model}
we will define precisely the contexts in which the agent asks this
question, including which variable $X$ it asks about.  The expert's
answer $X\in \Pi^+_Y$ gives priority to a variable $Y$ that she
believes the agent is unaware of. This increases the chances that the
agent will learn potentially valuable information about the hypothesis
space.

The expert's answer $X\in \Pi^+_Y$ has an observable interpretation;
$X\in \Pi^+_Y$ is added to $\delta_{0:t}$ and since $dn_t\models
\delta_{0:t}$, $X\in \Pi^t_Y$.  Thus some dependencies in $\Pi_t$ are
inferred monotonically via observable expert messages.  Others are
inferred defeasibly via statistical pattern recognition in the domain
trials (see Section~\ref{sec:model}).

\section{The Model for Learning}
\label{sec:model}

We now define \dn{} update in a way that meets the criteria from
Section~\ref{sec:task}.  We must estimate all components of the \dn{}
from the latest evidence $e_t$, in a way that satisfies the partial
description $\delta_{0:t}$ that has accumulated so far (this is
required to make the \dn{} valid).  That is, one must estimate the set
of random variables ${\cal C}_t$ and ${\cal A}_t$, the domain
$\Pi^t_R\subseteq {\cal C}_t$ of the reward function as well as the
function ${\cal R}_t$ itself, the dependencies $\Pi_t$ among ${\cal
  C}_t\cup {\cal A}_t$, and the conditional probability tables
(\cpt{}s) $\theta_t$, given $\Pi_t$.  These components are estimated
from the prior \dn{} $dn_{t-1}$, the latest evidence $e_t$ and the
partial description $\delta_{0:t}$ in the following order, as
shown in Figure~\ref{fig:architecture}:

\begin{enumerate}
	\item \begin{enumerate}
			\item  Estimate ${\cal B}_t$, ${\cal O}_t$, ${\cal A}_t$ and $\Pi^t_R$;
			\item Estimate ${\cal R}_t$, given $\Pi^t_R$;
		\end{enumerate}
	\item \begin{enumerate}
			\item Estimate $\Pi_t$, given ${\cal B}_t\cup {\cal O}_t\cup {\cal A}_t$ and $\Pi^t_R$;
			\item Estimate $\theta_t$, given $\Pi_t$.
		\end{enumerate}
\end{enumerate}

Step 1 proceeds via constraint solving (see
Section~\ref{sec:vocab-reward}); step 2 via a combination of symbolic
and statistical inference (see Section~\ref{sec:pi}).  Update proceeds
in this order because the set of dependencies one must deliberate over
depends on the set of random variables it is defined over, and the
global constraint on dependencies---namely, that the utility node is a
descendant of all nodes in the \dn---must be defined in terms of the
domain $\Pi^t_R$ of the reward function.  Likewise, the set of \cpt{}s
$\theta_t$ that one must estimate is defined by the dependency
structure $\Pi_t$.  The search for a {\em valid} \dn{} can prompt
backtracking, however: e.g., failure to derive a valid dependency
structure $\Pi_t$ may ultimately lead to a re-estimate of the set of
random variables ${\cal B}_t\cup {\cal O}_t\cup {\cal A}_t$.  We now
proceed to describe in detail each of these components of \dn{}
update.

\subsection{Random Variables and Reward Function}
\label{sec:vocab-reward}

The first step in \dn{} update is to identify $dn_t$'s random
variables---that is, the sets ${\cal B}_t$, ${\cal O}_t$ and ${\cal
A}_t$---and the reward function ${\cal R}_t$.  This is achieved via {\bf
constraint solving}, with the constraints provided by
$\delta_{0:t}$.

The number of valid vocabularies is always unbounded, because any
superset of a valid vocabulary is valid.  As motivated in
Section~\ref{sec:task}, we make search tractable via greedy search for
a {\em minimal} valid vocabulary: the agent (defeasibly) infers the
vocabulary in (\ref{eq:vocab}a--c) (this covers all the variables the
agent is aware of thanks to the dialogue strategy from
Section~\ref{sec:dialogue}), and also defeasibly infers the
minimal domain for ${\cal R}_t$, defined in
(\ref{eq:vocab}d).
\begin{subequations}
\label{eq:vocab}
\begin{align}
{\cal B}_t = & \{X: X\in {\cal B}_+\mbox{ is a conjunct in }
\delta_{0:t}\}\\
{\cal A}_t = & \{X: X\in {\cal A}_+\mbox{ is a conjunct in }
\delta_{0:t}\}\\
{\cal O}_t = & \{X: X\in {\cal O}_+\mbox{ is a conjunct in }
\delta_{0:t}\}\\
\Pi^t_R= & \{X: X\in \Pi^+_R \mbox{ is a conjunct in
}\delta_{0:t}\}
\end{align}
\end{subequations}

The evidence described in Section~\ref{sec:evidence} yields three
kinds of formulae in $\delta_{0:t}$ that (partially) describe ${\cal
R}_+$: (\ref{eq:valid-domain}), (\ref{eq:optimal-action-info}), and
$X\in \Pi^+_R$.  The agent uses constraint solving to find, or fail to
find, a reward function ${\cal R}_t$ that satisfies all conjuncts in
$\delta_{0:t}$ of the form (\ref{eq:valid-domain}) and
(\ref{eq:optimal-action-info}), plus the constraint
(\ref{eq:r-domain}), which will ensure ${\cal R}_t$ is well-defined
with respect to its (estimated) domain (\ref{eq:vocab}d).
\begin{equation}
\label{eq:r-domain}
\forall s_1\forall s_2((s_1\upharpoonright \Pi^t_R \leftrightarrow
s_2\upharpoonright \Pi^t_R) \rightarrow {\cal R}_t(s_1)={\cal R}_t(s_2))
\end{equation}

Constraints of the form (\ref{eq:valid-domain}) and
(\ref{eq:optimal-action-info}) are skolemized and fed into an
off-the-shelf constraint solver (with ${\cal R}_+$ replaced with
${\cal R}_t$).  The possible denotations of these skolem constants are
the atomic states defined by the vocabulary (\ref{eq:vocab}a--c).  If
there is a solution, the constraint solver returns for each skolem
constant a specific denotation $x\in v({\cal C}_t\times {\cal A}_t)$.

Substituting the skolem terms with their denotations projected onto $\Pi^t_R$ yields equalities ${\cal R}_t(y)=r$ (from (\ref{eq:valid-domain})) and inequalities ${\cal R}_t(y)>r$ (from (\ref{eq:optimal-action-info})), where $y\in v(\Pi^t_R)$.  A {\em complete} function ${\cal R}_t$ is constructed from this partial function by defaulting to indifference (recall Minimality): for any $y\in v(\Pi^t_R)$ where there is an inequality ${\cal R}_t(y)>r$ but no equality ${\cal R}_t(y)=r$, we set ${\cal R}_t(y)=r+c$ for some constant $c$ (in our experiments, $c=0.1$); for any $y\in v(\Pi^t_R)$ for which there are no equalities or inequalities, we set ${\cal R}_t(y)=0$.

The constraint solver may yield no solution: i.e., there is no function ${\cal R}_t$ with the currently estimated domain (\ref{eq:vocab}d) satisfying all observed evidence about states and their rewards.  This is the context ``unforeseen rewards'' described in Section~\ref{sec:dialogue} (see also Figure~\ref{fig:architecture}): the agent asks (\ref{eq:question-unforeseen-reward}), deferring \dn{} update until it receives the expert's answer.  The expert's answer is guaranteed to provide a new variable to add to $\Pi^t_R$.  It may also be a neologism---a variable the agent was unaware of---and so after updating $\delta_{0:t}$ with the expert's answer, the agent backtracks to re-compute (\ref{eq:vocab}a--d).

\subsection{Estimating $\langle \Pi_t,\theta_t\rangle$}
\label{sec:pi}
Current approaches to incrementally learning $\Pi$ in a graphical model of belief exploit {\em local} inference to make the task tractable.  There are essentially two forms of local inference.

The first is a greedy local search over a full structure: remove an
edge from the current \dag{} or add an edge that does not create a
cycle, then test whether the result has a higher likelihood given the
evidence (e.g., \namecite{bramley:etal:2015, friedman1997sequential}).
This is Conservative: $\Pi$ changes only when evidence justifies
it. However, adapting it to our task is problematic.  Firstly, such
techniques rely heavily on a decent initial \dag{} to avoid getting
stuck in a local maximum; but in our task the agent's initial
unawareness of the possibilities makes an initial decent \dag{} highly
unlikely. Secondly, removing an edge can break the global constraint
on \dns{} that all nodes connect to the utility node.  We would need
to add a third option of doing local search given $\Pi$: {\em
  replace} one edge in $\Pi$ with another edge somewhere else.  This
additional option expands the search space considerably.

We therefore adopt the alternative form of local inference: assume
conditional independence among parent sets.  \namecite{buntine:1991}
assumes that $X$'s parent set is conditionally independent of $Y$'s
given evidence $e_{0:t}$ and the total temporal order $\succ$ over the
random variables---$X\succ Y$ means that $X$ may be a parent to $Y$
but {\em not} vice versa.  This independence assumption on its own
is not sufficient for making reasoning tractable, however. If a Bayes
Net has 21 variables (as the ones we experiment with in
Section~\ref{sec:experiments} do), then a variable may have $2^{20}$
possible parent sets---a search space that is too large to be
manageable.  So \namecite{buntine:1991} prunes $X$'s possible parent
sets to those that evidence so far makes reasonably likely (we will
define ``reasonably likely'' in a precise way shortly). There are then
two alternative ways of updating:

\begin{itemize}
\item \textbf{Parameter Update}: Estimate the posterior probability of
  a parent set from its prior probability and the \emph{latest}
  evidence under an assumption that the set of reasonable parent sets
  \emph{does not change}.
\item \textbf{Structural Update}: Review and potentially \emph{revise}
  which parent sets are reasonable, given a {\em batch} of
  evidence. (Thus, Structural Update changes the set of possible
  structures which are considered).
\end{itemize} 

We adapt Buntine's model to our task in two ways. Firstly, Buntine's
model assumes that the total ordering on variables is known. In our
task the total order is hidden, and marginalising over all possible
temporal orders is not tractable---it is exponential on the size of the
vocabulary. We therefore make an even stronger initial independence
assumption than Buntine when deciding which parent sets are
reasonable: Specifically, that $X$'s parent set is conditionally
independent of $Y$'s given evidence $e_{0:t}$ alone.  Unfortunately,
this allows combinations of parent sets with non-zero probabilities to
be cyclic. We therefore have an additional step where we greedily
search over the \emph{space of total orderings} (similar to
\nncite{friedman2003being}), and use {\em Integer Linear Programming}
\nncite{vanderbei:2015} at each step to find a ``most likely'' causal
structure $\Pi_t$ which both obeys the currently proposed ordering and
that is also a valid \dn---in particular, it is a \dag{} where the
utility node is a descendant of all other nodes.

Secondly, as the agent's vocabulary of random variables expands, we need to provide new probability distributions over the larger set of possible parent sets, which in turn will get updated by subsequent evidence.  We now describe each of these components of the model in turn.

\subsubsection{Parameter Update}  Each variable $V\in {\cal C}_t$ is
associated with a set $P_v$ of \emph{reasonable parent sets}. Each
parent set $\Pi_v \in P_v$ is some combination of variables from
${\cal C}_t\cup {\cal A}_t$.  Parameter Update determines the
posterior distribution over $P_v$ given its prior distribution and the
latest piece of evidence under the assumption that the possible values
of $P_v$ do not change. 
	
\paragraph{Updates from Domain Trials}
Suppose that the latest evidence $e_t$ is a domain trial (i.e.,
$\tau_t\in\sample_{0:t}$). Parameter Update uses Dirichlet
distributions to support incremental learning: if
$\tau_t\upharpoonright V=i$ and $\tau_t\upharpoonright \Pi_v=j$, then
we can calculate the posterior probability of $\Pi_v$ in a single step
using (\ref{eq:corrected-parameter-update}):

\begin{equation}
\label{eq:corrected-parameter-update}
\begin{array}{l}
Pr(\Pi_v | e_{0:t}) 
= Pr(\Pi_v | e_{0:t-1})\frac{(n_{v=i|j} +
\alpha_{v=i|j} - 1)}{(n_{v=.|j} + \alpha_{v=.|j} - 1)}
\end{array}
\end{equation}

Here, $n_{v=i|j}$ is the number of trials in $\sample_{0:t}$ where $V=i$ and $\Pi_v=j$, and  $\alpha_{v=i|j}$ is a ``pseudo-count'' which represents the Dirichlet parameter for $V=i$ and $\Pi_v=j$. The sum of all trials where $\Pi_v=j$ is given by $n_{v=.|j}$.  Formula (\ref{eq:corrected-parameter-update}) follows from the recursive structure of the $\Gamma$ function in the Dirichlet distribution (the Appendix provides a derivation, which corrects an error in \nncite{buntine:1991}).

Estimating $\theta_t$, given $\Pi_t$, likewise exploits the Dirichlet distribution \nncite[p56]{buntine:1991}:

\begin{equation}
\label{eq:estimate-theta}
\displaystyle
\begin{array}{r l}
E_{\theta|\sample_{0:t},\Pi}(\theta_{v=i|j}) = & \displaystyle \frac{\int_{\theta}\theta_{v=i|j}Pr(\sample|\Pi,\theta)Pr(\theta|\Pi)}
{\int_{\theta}Pr(\sample|\Pi,\theta)Pr(\theta|\Pi)}\\
\\
= & \displaystyle \frac{n_{x=i|j}+\alpha_{v=i|j}}{n_{v=.|j} + \alpha_{v=.|j}}
\end{array}
\end{equation}

In words, (\ref{eq:estimate-theta}) computes the conditional probability tables (\cpt{}s) directly from the counts in the trials and the appropriate Dirichlet parameters, which in turn quantify the extent to which one should trust the counts in the domain trials for estimating likelihoods---the higher the value of the $\alpha$s relative to the $n$s, the less the counts influence the probabilities.  Note that the $\alpha$-parameters vary across the values of the variables and their (potential) parent sets.  We motivate this shortly, when we describe how to perform \dn{} update when a new random variable needs to be added to it.  At the start of the learning process, $\alpha_{v=i|j}=0.5$ for all $V$, $i$ and $j$.

\paragraph{Updates from Expert Evidence}

Now suppose $e_t$ is an {\em expert utterance} $\sigma_t$, but {\em not} one that introduces a neologism (we discuss \dn{} update over an expanded vocabulary at the end of this section).  Then Parameter Update starts by using $\wedge$-elimination on $\delta_t$ (i.e., the partial description of $dn_+$ that $\sigma_t$ entails) to infer a conjunction $\pi_{t,v}$ of all conjuncts in $\delta_{0:t}$ of the form $X\in \Pi_v$.  Note that $\delta_t$ does not contain conjuncts of the form $X\not\in \Pi_v$ (see Section~\ref{sec:dialogue}), although it may entail such formulae (e.g., from conjuncts declaring a variable's type).  Formula (\ref{eq:update-e-l}) then computes a posterior distribution over $P_v$, where $\eta$ is a normalising factor:

\begin{equation}
\label{eq:update-e-l}
Pr(\Pi_v | e_{0:t}) =
\left \{
\begin{array}{l l}
\eta Pr(\Pi_v|e_{0:t-1}) & \mbox{if } \Pi_v \vdash \pi_{t,v} \\
0 & \mbox{otherwise}
\end{array}
\right.
\end{equation}

Equation (\ref{eq:update-e-l}) is {\em Conservative}---it preserves the relative likelihood of parent sets that are consistent with $\pi_{t,v}$.  It offers a way to rapidly and incrementally update the probability distribution over $P_v$ with at least some of the information that is revealed by the latest expert evidence. 

\paragraph{Enforcing global constraints on $\Pi$}

Equations (\ref{eq:corrected-parameter-update}) and
(\ref{eq:update-e-l}) on their own do not comply with Satisfaction nor
even with Consistency. If we naively construct the ``most likely''
global structure by simply picking the highest probability parent set
$\Pi_v$ for each variable $V$ independently, their combination might
be cyclic, or violate information about parenthood entailed by
$\delta_{0:t}$.

To find a global structure which is \emph{valid} as well as likely, we combine ILP techniques with a greedy local search over the \emph{space of total orderings}.

We first compute $Pr(P_v|e_{0:t},\succ)$ from $Pr(P_v|e_{0:t})$, where
$\succ$ is a total temporal order over $dn_t$'s random variables that
satisfies the partial order entailed by $\delta_{0:t}$. Next, we use
ILP techniques to determine the most likely structure
$\Pi^{m}_{\succ}$ which obeys the current ordering $\succ$, and is a
\dag{} with all nodes connected to the utility node.  We stop
searching when all valid total orders $\succ'$ formed via a local
change to $\succ$ yield a less likely structure (i.e.,
$Pr(\Pi^{m}_{\succ'}|e_{0:t},\succ')\leq
Pr(\Pi^{m}_{\succ}|e_{0:t},\succ)$), and return $\Pi^{m}_{\succ}$
as $\Pi_t$. We now describe these steps in detail.

The partial description $\delta_{0:t}$ imposes a partial order on the
random variables, thanks to the expert's declarations of the form
$X\in \Pi_Y$ (this corresponds to condition (i) in
Fact~\ref{fact:total-order}), its entailments about each variable's
{\em type} (i.e., ${\cal A}_t$, ${\cal B}_t$ or ${\cal O}_t$), which
are in effect constraints on the relative position of variables within
the \dag{} (conditions (ii)--(iv) in Fact~\ref{fact:total-order}); and
information about whether a variable is an immediate parent to the
utility node (condition (v)):
\begin{fact}{\bf Total orders that satisfy $\delta_{0:t}$}\\
\label{fact:total-order}
A total order $\succ$ of random variables satisfies the partial order
entailed by
$\delta_{0:t}$ iff it satisfies the following 5 conditions:
\begin{enumerate}
\item   [(i)] If $\delta_{0:t}\models X\in \Pi_y$, then 
$X\succ Y$; 
\item   [(ii)] If $\delta_{0:t}\models A\in {\cal A}_+$ then
$X\not\succ A$ for any $X$, and $A\not\succ Y$ for any $Y$ where
$\delta_{0:t}\models Y\in {\cal B}_t$;
\item   [(iii)] If $\delta_{0:t}\models O\in {\cal O}_+$ then there is
a variable $X$ where $\delta_{0:t}\models X\in {\cal A}_+$ and
$X\succ O$;
\item   [(iv)] If $\delta_{0:t}\models B\in {\cal
B}_+ \wedge O\in {\cal O}_+$ then $B\succ O$; and 
\item   [(v)]
for any $X,Y$, if $\delta_{0:t}\models X\in \Pi^+_R$ and
$\delta_{0:t}\not\models Y\in \Pi^+_R$, then $Y\succ X$.
\end{enumerate}
Note that thanks to $dn_0$ and the dialogue strategies, where $\Phi$
is ${\cal B}_+$, ${\cal O}_+$, ${\cal A}_+$, $\Pi^+_R$ or $\Pi_y$, one
can test whether $\delta_{0:t}\models X\in \Phi$ via
$\wedge$-elimination.
\end{fact}

The agent starts by choosing (at random) a total order $\succ$ that
satisfies Fact~\ref{fact:total-order}.  The agent then uses
(\ref{eq:parent-succ}) to estimate the probabilities over direct
parenthood relations ($pa(X,Y)$ means $X$ is a parent to $Y$) given
the evidence and $\succ$:
\begin{equation}
\label{eq:parent-succ}
\begin{array}{r l}
Pr(pa(X,Y)|e_{0:t},\succ) = & 
\left\{
\begin{array}{l l}
\sum_{\Pi_y\in P_y:X\in \Pi_y}
Pr(\Pi_y|e_{0:t}) & \mbox{if } X\succ Y\\
0 & \mbox{if } Y\succ X
\end{array}
\right.
\end{array}
\end{equation}
Thanks to Fact~\ref{fact:total-order} this makes any combination of
non-zero probability parenthood relations a \dag{} with no ${\cal
B}$-variable being a descendant to an ${\cal A}$ variable, and all
${\cal A}$ variables have no parents.  But it does not guarantee that
all nodes are connected to the utility node.  We use an ILP step to
impose this global constraint.  The result is a valid structure $\Pi$
(i.e., it satisfies $\delta_{0:t}$) that evidence so far also deems to
be likely (we give an outline proof of its validity on
page~\pageref{fact:valid}).

The {\bf ILP Step} is formally defined as follows:
\begin{description}
\item 
[Decision variables:] Where $X,Y\in {\cal C}_t\cup {\cal
A}_t$, $\mbox{{\em 
pa}}(X,Y)$ is a Boolean variable with value 1
if $X$ is a parent of $Y$ (i.e., $X\in \Pi_y$),
and 0 otherwise 
\item   [Objective Function:]  We want to find the most likely
combination of valid
parenthood relations: i.e., 
we want to solve (\ref{eq:objective}):
\begin{equation}
\label{eq:objective}
max \sum_{X,Y \in \mathcal{C}_t \cup \mathcal{A}_t} Pr(pa(X,Y)) * pa(X,Y) + (1 - Pr(pa(X,Y))) * (1 - pa(X,Y))
\end{equation}
\item [Constraints:]  C1 ensures three things: (i) variables
cannot have a parent which is not present in at least one of
the reasonable parent sets (see (\ref{eq:parent-succ})); (ii)
parent sets obey all expert declarations of the form
$Z\in \Pi_{z'}$ (by
equation (\ref{eq:update-e-l})); and (iii) parents obey the
currently proposed temporal ordering $\succ$ (by equation
(\ref{eq:parent-succ})) thereby guaranteeing that the final
graph is 
acyclic with no ${\cal B}$ variable being a descendant of
${\cal A}$ and all ${\cal A}$ variables are orphans. Constraints
C2 and C3 ensure that the final 
graph satisfies the necessary global conditions for being
a \dn{}, namely that every variable is in some way
connected to the utility node, and that every outcome
variable has at least one action ancestor. C4 ensures that
the decision variables take on binary values (0 or 1). 
\[
\begin{array}{l l r}
C1:\quad &\forall X,Y. (Pr(pa(X,Y)) = 0 \implies pa(X,Y) = 0) &\text{Disallow impossible parents} \\
C2:\quad &\forall X \notin \Pi^t_R. (\sum_{y} pa(X,Y) \geq 1) & \text{One child minimum}
\\
C3:\quad &\forall O \in \mathcal{O}. (\sum_{y \in \mathcal{O}
\cup \mathcal{A}} pa(Y, O)\geq 1) & \text{No orphaned outcomes} \\
C4:\quad &\forall X,Y. pa(X,Y) \in \{0,1\} &\text{Binary
Restriction}
\end{array}
\]
\end{description}

\paragraph{Linear Relaxation of the ILP Step}
Finding the solution to a Linear Program where all decision variables
must be integers is an NP-hard problem. So we approximate the solution
via \emph{Linear Relaxation}: we replace C4 with (\ref{eq:c5lp}), so
that the values of decision variables can be real numbers:
\begin{equation}
\label{eq:c5lp}
\forall X,Y. pa(X,Y) \in [0,1]
\end{equation}
We then round the resulting values to integers.  This reduces the time
needed to compute a solution, but one must be careful in the rounding
procedure: simply rounding each $pa^*(X,Y)$ to its nearest integer may
create an inconsistent \dn{}.  Rounding in two phases avoids this.  In
the first phase, we produce a set of rounded decision variables
$pa'(X,Y)$ that satisfy C1 and C2:
\[
\begin{array}{l l}
pa'(X,Y) = 
\left \{
\begin{array}{l l}
1 &  \mbox{if }
pa^*(X,Y) \geq 0.5 \vee 
(X \notin \Pi^t_R \wedge \forall Z \neq Y. (pa^*(X,Y) \geq pa^*(X,Z)))\\
0 & \mbox{otherwise}
\end{array}
\right.
\end{array}
\]
In the second phase, we ensure that constraint C3 is also satisfied, 
producing the final integer values for all the variables $pa(X,Y)$:
\[
\begin{array}{l}
pa(X,Y) = 1 \mbox{ iff } pa'(X,Y) = 1 \quad \vee \\
\qquad
(Y \in \mathcal{O} \wedge\\
\qquad \qquad \forall Z((Z \neq X \wedge Z \in \mathcal{O}
\cup \mathcal{A}) \rightarrow\\
\qquad\qquad\qquad (pa'(Z,Y) = 0 \wedge Pr(pa(X,Y)) >
Pr(pa(Z, Y)))))
\end{array}
\]

Because the random variable $P_v$ does not contain {\em all} possible
parent sets to $V$ (so as to keep learning tractable), it is possible
that there is in fact no solution to the ILP step for any valid
temporal order $\succ$.  This is a context in which the agent performs
a Structural Update, which in turn changes the variables that satisfy
the antecedent to constraint C1.

\subsubsection{Structural Update}
\label{subsec:structural_update}

To make learning tractable, $P_v$ includes only reasonable parent sets
to $V$---a strict subset of those that are possible.  But as evidence
changes, so does what is reasonably likely.  Therefore, the agent
occasionally performs a {\em Structural Update}: the full batch of
evidence $e_{0:t}$ is used to review and potentially revise the
possible values of $P_v$ and its probability distribution.

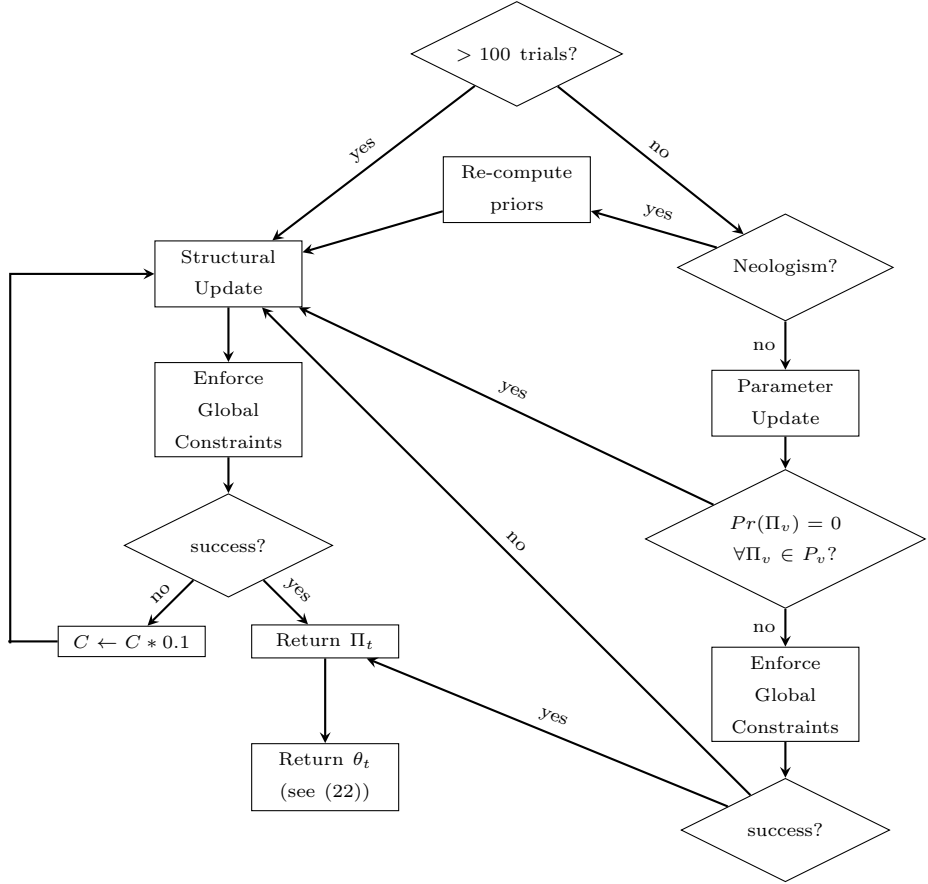
\begin{figure}[t]
\centering
\tikzstyle{process} = [rectangle, aspect=1, text centered, draw=black, text width=1.7cm]
\tikzstyle{decision} = [diamond, aspect=2, text width=1.7cm, text centered, draw=black]
\tikzstyle{arrow} = [thick,->,>=stealth]
\tikzstyle{edge} = [thick,-,>=stealth]

\begin{tikzpicture}[node distance=1.8cm]
\node(100)[decision]{\scriptsize{$>100$ trials?}};
\node(su)[process, below left = 3cm of 100]{\scriptsize{Structural Update}};
\node(neo)[decision, below right = 3cm of 100]{\scriptsize{Neologism?}};
\node(neo-yes)[process, below of=100]{\scriptsize{Re-compute priors}};
\node(pu)[process, below of=neo]{\scriptsize{Parameter Update}};
\node(egs-su)[process, below of=su]{\scriptsize{Enforce Global
Constraints}};
\node(egs-su-?)[decision, below of=egs-su]{\scriptsize{success?}};
\node(egs-su-yes)[process, below right of=egs-su-?]{\scriptsize{Return
$\Pi_t$}};
\node(su-theta)[process,below of=egs-su-yes]{\scriptsize{Return $\theta_t$
(see (\ref{eq:estimate-theta}))}};
\node(egs-su-no)[process, below left of=egs-su-?]{\scriptsize{$C\leftarrow
C*0.1$}};
\node(dummy1)[left=0.5cm of egs-su-no]{};
\node(pu-pv)[decision,below of=pu]{\scriptsize{$Pr(\Pi_v)=0$ $\forall\Pi_v\in P_v$?}};
\node(egs-pu)[process, below=0.5cm of pu-pv]{\scriptsize{Enforce Global
Constraints}};
\node(egs-pu-?)[decision, below of=egs-pu]{\scriptsize{success?}};
\node(arrow-point)[left=9cm of egs-pu-?]{};
\draw[arrow](100) -- node[sloped,above]{\scriptsize{yes}} (su);
\draw[arrow](100) -- node[sloped,above]{\scriptsize{no}} (neo);
\draw[arrow](neo) -- node[sloped,above]{\scriptsize{yes}} (neo-yes);
\draw[arrow](neo-yes) -- (su);
\draw[arrow](neo) -- node[anchor=east]{\scriptsize{no}} (pu);
\draw[arrow](su) -- (egs-su);
\draw[arrow](egs-su) -- (egs-su-?);
\draw[arrow](egs-su-?) -- node[sloped,above]{\scriptsize{yes}} (egs-su-yes);
\draw[arrow](egs-su-yes) -- (su-theta);
\draw[arrow](egs-su-?) -- node[sloped,above]{\scriptsize{no}} (egs-su-no);
\draw[edge,shorten >=-1.5mm](egs-su-no) -- (dummy1);
\draw[arrow,shorten <=-1.5mm](dummy1) |- (su);
\draw[arrow](pu) -- (pu-pv);
\draw[arrow](pu-pv) -- node[anchor=east]{\scriptsize{no}} (egs-pu);
\draw[arrow](pu-pv) -- node[sloped,above]{\scriptsize{yes}} (su);
\draw[arrow](egs-pu) -- (egs-pu-?);
\draw[arrow](egs-pu-?) -- node[sloped,above]{\scriptsize{yes}} (egs-su-yes);
\draw[arrow](egs-pu-?) -- node[sloped,above]{\scriptsize{no}} (su);
\end{tikzpicture}

\caption{Algorithm for Estimating $\langle\Pi_t,\theta_t\rangle$.}
\label{fig:flow-chart}
\end{figure}

Figure~\ref{fig:flow-chart} depicts the structure of the algorithm
that corresponds roughly to the nodes ``estimate $\Pi_t$'' and
``estimate $\theta_t$'' in Figure~\ref{fig:architecture}.  In
particular, it depicts the four contexts in which Structural Update is
performed in our experiments in Section~\ref{sec:experiments}:

\begin{enumerate}
\item There have been 100 domain trials since last structural update
\item The agent has just discovered an unforeseen factor
\item The latest dialogue evidence $\sigma_t$ yields a 0 probability to all parent sets in $P_v$ via equation (\ref{eq:update-e-l})
\item The algorithm for enforcing global constraints on $\Pi$ fails to produce any valid structure
\end{enumerate} 

The {\bf input} to Structural Update consists of a (chance) variable
$V$, the batch of evidence $e_{0:t}$, the prior distribution
$Pr_{t-n}(\Pi_v)$ over {\em all} possible parent sets to $V$ (where
$t-n$ is the time of the previous structural update), and the current
partial description $\delta_{0:t}$ of the \dn.  Its {\bf output} is
a (perhaps new) set of values for $P_v$ (i.e., those parent sets to
$V$ that are currently deemed ``reasonable''), the posterior probability
$Pr_t(\Pi_v)$ for each possible parent set $\Pi_v$ (whether it is in
$P_v$ or not) and the posterior probability distribution $Pr_t(P_v)$
(which is computed by normalising $Pr_t(\Pi_v)$ for $\Pi_v\in P_v$).
Informally, the Structural Update {\bf algorithm} dynamically
constructs a {\em parent lattice} $L^t_v$, with each node
corresponding to a parent set to $V$ and arcs corresponding to the
{\em superset} relation.  When expanding the lattice by adding a new
node, we estimate its posterior probability given the evidence
$e_{0:t}$.  We stop expanding the lattice when all its leaves have a
posterior probability that is below a certain threshold.  Thus
Minimality is implicit: Structural Update assumes that supersets of
sufficiently improbable parent sets will not improve their probability.

More formally, the root(s) of the parent lattice $L^t_v$ are the
minimal valid parent sets: i.e., they contain $X$ if
$\delta_{0:t}\models X\in \Pi_v$, and if $\delta_{0:t}\models V\in
{\cal O}_t$ then $L^t_v$'s root(s) each contain at least one variable
from ${\cal A}_t\cup{\cal O}_t$.  For each root $\Pi_v$, one computes
its posterior probability, given the evidence $e_{0:t}$.  If a node
was alive in the previous Structural Update, then this posterior
probability is already calculated via the sequence of Parameter
Updates that were performed between the previous and current
structural updates. It is therefore immediately available. If the node
was marked as asleep in the previous lattice, however, then the node's
posterior probability will not have been updated with the intervening
domain trials---it is computed now by performing a sequence of
Parameter Updates (see equation
(\ref{eq:corrected-parameter-update})).

The highest of these posterior probabilities is taken to be the {\em
Best-Posterior}.  One then marks the nodes as {\em alive} if its
posterior probability is $\geq C * \mbox{{\em Best-Posterior}}$ for some
constant $C$ (in the experiments in Section~\ref{sec:experiments},
$C=0.001$), and {\em asleep} otherwise.  You then create a new node in
the lattice by adding a variable to an alive leaf node.  You compute
its posterior probability, and if it exceeds the current {\em
Best-Posterior}, then you update {\em Best-Posterior} and
re-classify which nodes in the lattice are alive and which are asleep
accordingly.  You stop when all the leaves in the current lattice are
asleep.

The set $P_v$ is defined to be all the alive parent sets in $L^t_v$
and their ancestors in $L^t_v$.\footnote{This contrasts with Buntine's
Structural Update, in which $P_v$ is the alive parent sets {\em
only}.  We include asleep subsets of alive parent sets because our
version of Parameter Update abstracts over all possible total
temporal orders of the variables.  To illustrate the issue, suppose
that there is a very strong statistical correlation in the domain
trials between $X$ and $Y$---so strong that the only alive parent
set for $P_x$ is $\{Y\}$, and the only alive parent set for $P_y$ is
$\{X\}$.  If $P_x=\{\{Y\}\}$ and $P_y=\{\{X\}\}$, then these
generate no consistent structure.  But if $P_v$ includes all subsets of
alive parent sets, then we avoid this problem: $\emptyset$ will be a
possible value of $P_x$ and of $P_y$, even though
Parameter Update deems $\emptyset$ to be an unlikely parent set for
$X$ and for $Y$.}  Having (re-)estimated $P_v$ for each variable
$V$, you {\em test} whether each variable $X\not\in\Pi^t_R$ is in at
least one reasonable parent set: if it is not then there is clearly no
valid $\Pi$ that one can generate from them (since $X$ will not connect to
the utility node).  This test failing
is the context ``Unknown Effects'' in the agent's
dialogue strategy, which we described in
Section~\ref{sec:dialogue:agent}: i.e., instead of expanding the
search space by reducing the asleep threshold $C$, the agent seeks
further evidence by asking the expert (\ref{eq:which-effect}): what is
an effect of $X$?  The evidence $e_{0:t}$ and partial description
$\delta_{0:t}$ get updated with the expert's answer, and \dn{} update
begins again---backtracking to estimating the random variables via
(\ref{eq:vocab}a--c) may be necessary because the expert's answer may be a
neologism.  The experiments in Section~\ref{sec:experiments} show that
this querying strategy keeps a tight reign on the size of the search
space, while still ensuring that many of the dependencies in $\Pi_t$
are inferred defeasibly from the domain trials, rather than
monotonically from explicit expert declarations about parenthood.

The posterior probabilities that are computed during the lattice
expansion are assumed to be the posterior probabilities
$Pr_t(\Pi_v|L^t_v)$---the probability that $\Pi_v$ is $\Pi^+_v$, given
that $\Pi^+_v\in L^t_v$ (if $\Pi_v\not\in L^t_v$, then
$Pr_t(\Pi_v|L^t_v)=0$).  The posterior probabilities on {\em every
possible} parent structure is then defined by
(\ref{eq:poss-parent-set}) to (\ref{eq:prob-lattice}), where
$Pr(L^t_v)$ is the probability that the true parent set $\Pi^+_v$ is
in $L^t_v$, and $p$ is the probability mass we wish to assign to the
unexplored space:
\begin{equation}
\label{eq:poss-parent-set}
Pr_t(\Pi_v) = Pr_t(L_v^t)Pr_t(\Pi_v | L_v^t) + 
Pr_t(\neg L_v^t)Pr_t(\Pi_v | \neg L_v^t) 
\end{equation}
\begin{equation}
\label{eq:unreasonable-parent-set}
\begin{array}{r l}
Pr_t(\Pi_v | \neg L^t_v) = &
\left \{
\begin{array}{l l}
0 & \mbox{if } \Pi_v\in L^t_v\\
\dfrac{1}{2^{|\mathcal{C}_t \cup \mathcal{A}_t \setminus V|} - |L_v^t|} &
\mbox{if } \Pi_v \not\in L_v^t
\end{array}
\right.
\end{array}
\end{equation}
\begin{equation}
\label{eq:prob-lattice}
\begin{array}{r l}
Pr_t(\neg L_v) = & p \\
Pr_t(L_v) = & 1 - Pr_t(\neg L_v)
\end{array}
\end{equation}
Thus Structural Update returns for each chance variable $V$ the values
$P_v$, $Pr_t(P_v)$ and $Pr_t(\Pi_v)$ for all possible parent sets
$\Pi_v$.  But the algorithm does not require the agent to enumerate or
reason about parent sets within the unexplored space.  These become
relevant only if in the next Structural Update, the agent expands the
parent lattice to include a parent set that was absent from the prior
lattice.

The process for estimating $\Pi_t$ concludes, as always, by applying
the algorithm for enforcing global constraints that we described
earlier (see Figure~\ref{fig:flow-chart}). If this succeeds, it
returns $\Pi_t$, and from this one computes $\theta_t$ via
(\ref{eq:estimate-theta}).  If not, then the agent reduces the asleep
threshold (by multiplying $C$ by 0.1) and attempts Structural Update
again.

\subsubsection{Initial Distribution}

We implicitly encode the defeasible principle of Minimality in our
initial distribution over parent sets by making larger parent sets
less likely than smaller ones. The agent starts with a
small probability $\rho$ that $Y$ is a parent to $X$ (provided this is
consistent with the variables' types as stipulated in $\delta_0$). The
probability of a parent set $\Pi_x$ is then the product of each
individual parent's presence or absence:

\begin{equation}
\label{eq:prior-parenthoods}
\begin{array}{r l}
Pr_0(pa(Y,X)) \propto  &
\left\{
\begin{array}{l l}
0 & \mbox{if } \delta_0 \models \neg pa(Y,X)\\
\rho & \mbox{otherwise}
\end{array}
\right.
\end{array}
\end{equation}
\begin{equation}
\label{eq:prior-parent-sets}
\begin{array}{r l}
Pr(\Pi_x) = & \displaystyle\prod_{Y\in \Pi_X}Pr(pa(Y,X))\displaystyle\prod_{Z\not\in \Pi_X}(1-Pr(pa(Z,X)))
\end{array}
\end{equation}
For each $X\in {\cal C}_0$, the agent uses
(\ref{eq:prior-parenthoods}), (\ref{eq:prior-parent-sets}) and
Structural Update to dynamically construct the set $P^0_x$ of
reasonable parent sets of $X$ and its initial probability
distribution.

\subsubsection{Updating with Evidence containing a Neologism}

Suppose the latest evidence $e_t$ is an expert utterance $\sigma_t$
that features a {\em neologism} $Z$.  Then by (\ref{eq:vocab}a--c), $Z$ is
a random variable in $dn_t$ but it was not a part of $dn_{t-1}$.  This
means that the prior distributions over the possible parent sets no
longer cover all possible parent sets!  These must be updated to take
account of the extra possibilities afforded by the addition of $Z$,
and to ensure that the definitions we have given so far for calculating
posterior distributions from evidence (e.g., equation
(\ref{eq:corrected-parameter-update})) remain well-defined.  In
addition, the agent cannot observe $Z$'s past values in
$\sample_{0:t}$---i.e., it cannot observe $n_{z=i|j}$, nor $n_{v=i|j}$
when $Z\in \Pi_v$.  However, we need these counts for Parameter Update (see
(\ref{eq:corrected-parameter-update})) and estimating $\theta_t$
(see (\ref{eq:estimate-theta})).  The Dirichlet $\alpha$-parameters in
(\ref{eq:corrected-parameter-update}) and (\ref{eq:estimate-theta})
that involve $Z$ are also not defined.  We now describe how the agent
revises all these parameters, so as to ensure that
(\ref{eq:corrected-parameter-update}) and (\ref{eq:estimate-theta})
remain well defined and continue to support probabilistic
reasoning over the newly expanded hypothesis space.

We start by defining how the agent computes the probability
distribution over the expanded set of possible parent sets.  The agent
starts this process by performing a {\em Structural Update} on the
batch of evidence $e_{0:t-1}$ that preceded the latest evidence $e_t$
with the neologism $Z$. This ensures that the (small) probability mass
$p$ that is assigned to the set of ``unreasonable'' parent sets takes
into account all evidence to date.  This yields a revised (prior)
probability distribution $Pr(\Pi_x|e_{0:t-1})$ for every possible
parent set $\Pi_x$ of $X\in {\cal C}_t$ that does {\em not} include
the new variable $Z$.  Equation
(\ref{eq:old-parent-sets-new-variable}) then assigns probabilities to
{\em all} possible parent sets to $X\neq Z$, including parent sets
containing $Z$:
\begin{equation}
\label{eq:old-parent-sets-new-variable}
Pr(\Pi_x | e_{0:t}) \propto 
\left \{
\begin{array}{l l}
0 & \mbox{if } \delta_{0:t} \models \neg \Pi_x \\
(1-\rho) Pr(\Pi_x | e_{0:t-1})  & \mbox{if } Z\not\in\Pi_x \\
\rho Pr(\Pi'_x | e_{0:t-1}) & \mbox{if } \Pi_x = \Pi'_x\cup 
\{Z\}
\end{array}
\right.
\end{equation}

This update is Conservative, because it preserves the relative
likelihood among the parent sets that do not include $Z$---in
particular, unreasonable parent sets remain unreasonable.  It is also
Minimal because it re-assigns only a small proportion $\rho$ of the
probability mass of a parent set that does not include $Z$ to the
parent set formed by adding $Z$ to it---in particular, adding $Z$ to
an unreasonable parent set is unreasonable.  As usual, the agent
needn't actively enumerate each possible parent set nor compute its
probability; rather (\ref{eq:old-parent-sets-new-variable}) is used to
dynamically construct the lattice $L^t_x$, from which the agent
identifies the reasonable parent sets $P^t_x$ for $X$ and $P^t_x$'s
probability distribution, where the possibilities now include the
additional variable $Z$.

If $Z$ is a chance variable, then equation
(\ref{eq:parenthood-new-variable}) in combination with
(\ref{eq:prior-parent-sets}) defines $Pr(\Pi_z|e_{0:t})$ (if $Z$ is an
action variable, then $\Pi_Z=\emptyset$):
\begin{equation}
\label{eq:parenthood-new-variable}
\begin{array}{r l}
Pr(pa(Y,Z)|e_{0:t}) = &
\left \{
\begin{array}{l l}
1 \qquad &\text{if } \delta_{0:t} \models pa(Y,Z)\\
0 \qquad &\text{if } \delta_{0:t}\models \neg pa(Y,Z)\\
\rho \qquad &\text{otherwise}
\end{array}
\right.
\end{array}
\end{equation}
This is Minimal because any variable being a parent to $Z$ is assigned
a low default probability $\rho$; so the dynamic lattice construction
$L^t_z$ for identifying $P^t_z$ and its probability distribution will
restrict search considerably.  With $P_v$ and its distribution for all
$V\in {\cal C}_t$ in place, one applies the algorithm for enforcing
global constraints to yield $\Pi_t$.

We now return to the issue of the counts $n_{v=i|j}$, and the
Dirichlet $\alpha$-parameters.  Our model avoids the computational
complexity of marginalising over $Z$'s past values by {\em throwing
away} the past domain trials $\sample_{0:t-1}$ and their counts
$n_{v=i|j}$, but {\em retaining} the relative likelihoods they gave
rise to by packing these into updated values for the Dirichlet
$\alpha$-parameters, as shown in (\ref{eq:alpha}) (where $K$ is a
constant):
\begin{equation}
\label{eq:alpha}
\begin{array}{l}
\mbox{Where $Z\in \dn_t$ but $Z\not\in \dn_{t-1}$}\\
\begin{array}{r l}
n_{v=i|j} = & 0 \mbox{ for all $v$, $\Pi_v$}\\
\alpha_{v=i|j} = & 
\left \{
\begin{array}{l l}
K*Pr_{t-1}(V=i,\Pi_v=j) & \mbox{ if $V\neq Z$ and $Z\not\in\Pi_v$}\\
K*0.5 & \mbox{ if $V=Z$ or $Z\in \Pi_v$}
\end{array}
\right.
\end{array}
\end{array}
\end{equation}
With the $\alpha$-parameters and the counts $n_{v=i|j}$ now defined
over the expanded set of atomic states, the agent can compute $\theta_t$, given
$\Pi_t$ and $e_{0:t}$, using (\ref{eq:estimate-theta}).  Furthermore,
it ensures that the equations (\ref{eq:corrected-parameter-update})
and (\ref{eq:estimate-theta}) remain well-defined, should the \dn{}
get updated by {\em subsequent} observed evidence.

Equation (\ref{eq:alpha}) makes the agent {\em Conservative} because when it observes subsequent evidence, the revised $\alpha$-parameters ensure that the likelihoods inferred from the past domain trials bias the estimated likelihoods of $\Pi$ and of $\theta$ that are based on that subsequent evidence (via equations (\ref{eq:corrected-parameter-update}) and (\ref{eq:estimate-theta}) respectively). Indeed, the larger $K$ is, the more Conservative you are: i.e., the more the probability distribution over dependencies and \cpt{}s before your set of random variables changed influences reasoning about dependencies and \cpt{}s after the set of random
variables changes.

\subsubsection{DN Update is Valid}

We have now defined how to use evidence to update each of the \dn's
components.  It meets the desiderata from Section~\ref{sec:task}.  In
particular:
\begin{fact}
\label{fact:valid}
\dn{} update complies with Consistency and Satisfaction: i.e., $dn_t$
is a consistent \dn{}, and $dn_t\models \delta_{0:t}$. 
\end{fact} 

\paragraph{Outline proof}
$\delta_{0:t}$ contains three kinds of conjuncts: (i) $X\in {\cal
B}_+$, $X\in {\cal A}_+$, $X\in {\cal O}_+$, $X\in \Pi^+_R$; (ii)
formulae of the form (\ref{eq:valid-domain}) and
(\ref{eq:optimal-action-info}); and (iii) $X\in \Pi_y$.  Equations
(\ref{eq:vocab}) together with the
algorithm for enforcing global constraints guarantees that $\Pi_t$
satisfies all the conjuncts of type (i), including their consequences
on each variable's relative position in the causal structure $\Pi^+$.
This is because $\Pi_t$ satisfies a total order $\succ$ (see equation
(\ref{eq:parent-succ}) and C1 in the ILP step), where by
Fact~\ref{fact:total-order} $\succ$ satisfies the partial order
imposed by $\delta_{0:t}$---e.g., where $\delta_{0:t}\models X\in
{\cal B}_+$, $\Pi^t_x\subset {\cal B}_t$.  ${\cal R}_t$ is a
consistent preference function because: (a) equation
(\ref{eq:r-domain}) makes ${\cal R}_t$ well-defined on its domain
$\Pi^t_R$; and (b) since its range is $\mathbb{R}$ it defines an
asymmetric and transitive relation.  Further, since all conjuncts in
$\delta_{0:t}$ of form (ii) are constraints on ${\cal R}_+$, it
follows immediately that the constraint solver returns a reward
function ${\cal R}_t$ that satisfies these.  Equation
(\ref{eq:update-e-l}) plus C1 in the ILP step guarantees that $\Pi_t$
entails conjuncts of type (iii).  Furthermore, the algorithm for
enforcing global constraints guarantees $\Pi_t$ is a \dag{} with all
nodes connected to the utility node: it is a \dag{} because $\Pi_t$
satisfies a consistent total order $\succ$; and all nodes are
connected to the utility node because of constraints C2 and C3 in the
ILP Step.  Finally, equation (\ref{eq:estimate-theta}) guarantees that
$\theta_t$ is a consistent probability distribution.  Thus $dn_t$ is
consistent, and $dn_t\models \delta_{0:t}$. 
$\blacksquare$

\section{Experiments}
\label{sec:experiments}

Our experiments show that our agent can learn the optimal policy for a
given task, even when initially unaware of factors that are critical
to its success. Further, we show the usefulness of the defeasible
principles our agent adheres to (such as Minimality and
Conservativity) by comparing our agent with several variations which
abandon those principles.

We evaluate our agent against two different scenarios. In the first
scenario, we have the agent learn in 10 randomly generated decision
problems. By varying the true decision problem the agent must learn,
these experiments evaluate whether the model's ability to converge on
optimal behaviour is robust to the type of domain it has to learn
about. In the second scenario, we have the agent learn the
hand-crafted ``Barley'' example from the introduction (see
Figure~\ref{fig:barley}). This experiment allows us to show how the
model performs on a concrete example of our novel task which is
explicitly designed to require the exploitation of unforeseen
possibilities to be successful.

Our primary metric is {\bf policy error} (PE). Equation
(\ref{eq:policy-error}) defines $PE_t$ at time $t$ as the weighted
average of the expected difference in reward between the true optimal
policy $\pi^*_+$ and the agent's perceived optimal policy $\pi^*_t$
(observations $b$ are projected onto the agent's current
conceptualisation of the domain to ensure $\pi^*_t$ is defined):

\begin{equation} 
	\label{eq:policy-error} 
	PE_t = \displaystyle\sum_{b\in v({\cal B}_+)} Pr_+(b) \left[EU_+(\pi^*_+(b)|b) - EU_+(\pi^*_t(b\upharpoonright {\cal B}_t)|b) \right]
\end{equation}

We also measure the agent's {\bf cumulative reward} over the learning period:

\begin{equation} 
	\label{eq:cum-reward}
	R_t=\displaystyle\sum_{\tau_i\in \sample_{0:t}} r_i
\end{equation}

Additionally, we measure the \textbf{runtime} the agent takes to
reason with the 3000 training samples. We do not measure the agent's
performance in terms of, for instance, the minimum edit distance
between $dn_t$ and $dn_+$, because the mapping from an optimal policy
to its \dn{} is one to many---for us, constructing a \dn{} is merely
the means for achieving optimal behaviour, and structural differences
between $dn_+$ and $dn_t$ may be benign in this respect.

All experiments are run over 3000 training examples. However, we
assume that each time the expert sends a message, the agent misses out
on an opportunity to interact with the domain. This means that the
more the expert intervenes, the fewer domain trials the agent gets
rewards from.


\subsection{Random \dns{}: Setup}

In these experiments, we tested each agent variation against 10
different decision problems. Each true decision network $dn_+$
consists of 21 Boolean variables: 7 action variables; 7 ``before''
variables; and 7 ``outcome'' variables (i.e., $|{\cal A}_+|=|{\cal
  B}_+|=|{\cal O}_+|=7$).  This results in non-trivial decision problems
with over 2 million atomic states.  We generate both the
structure and parameters for each $dn_+$ randomly using an adapted
version of the Bayesian network generation algorithm in
\nncite{ide2004generating}.\footnote{Our adaptation adds additional
  rules to ensure the algorithm generates only valid \dns{}. For
  example, the structure must conform to the rules of each variable's
  type.} We also create a reward domain for each $dn_+$ by choosing 5
chance variables at random, then generate a reward function which
yields rewards in the range 0--50.

We then assign an agent an initial conceptualisation $dn_0$ of the
decision problem; we test how unawareness affects learning by varying
$dn_0$.  To evaluate an agent's performance in learning, we run 100
simulations, each consisting of the agent observing and learning from
3000 pieces of evidence (a mix of domain trials and expert messages).
Running 100 simulations allows us to smooth over the fact that
learning may be affected by the non-deterministic outcomes of the
actions it performs.  Our performance metrics then take the average
over the 100 simulations.

We evaluate and compare agents that vary along the following dimensions:

\begin{description}
\item [Initial Awareness:] We compare an agent which starts at $dn_0$
with full knowledge of the (true) decision problem's set of random
variables against one
which starts with just one action and outcome variable:

\begin{center}
\begin{tabular}{@{}l  l@{}}
& Vocab Size at $dn_0$ \\ \midrule
Min-Vocab & 2\\
Full-Vocab & 21
\end{tabular}
\end{center}

\item [Expert Tolerance:] We vary the expert's tolerance to the
agent's suboptimal behaviour when she deliberates over whether to
offer unsolicited advice: this is achieved by varying $\beta$ and
$\gamma$ in (\ref{eq:condition-for-advice-proportion}):
\begin{center}
\begin{tabular}{@{}lll@{}}
& $\beta$ & $\gamma$ \\ \midrule
Low-Tolerance & 0.001 & 1 \\
Default-Tolerance & 0.9 & 50
\end{tabular}
\end{center}
\item [Minimality:] An agent can be minimal, slightly minimal, or
maximal.  They vary on the following three dimensions: (i) adopting
a greedy search for a {\em minimal} domain of ${\cal R}$ (see
equation (\ref{eq:vocab}d)) vs.\ defining the domain of
${\cal R}_t$ to be all of ${\cal C}_t$ (i.e., the
maximum possible domain, creating many more arrows in the graphical
component of the \dn); (ii) the prior probability $\rho$ that is
assigned to the parenthood relations (see equations
(\ref{eq:prior-parenthoods}) and
(\ref{eq:old-parent-sets-new-variable})) can make parenthood
(defeasibly) unlikely ($\rho=0.1$) or equally likely as unlikely
($\rho=0.5$); and (c) the asleep threshold $C$ that prunes the
search space to reasonable parent sets can result in pruning
($C=0.001$) or not ($C=0$).

\begin{center}
\begin{tabular}{@{}llll@{}}
& Minimal $\Pi^t_R$? & $\rho$ & $C$ \\ \midrule
Minimal & yes & 0.1 & 0.001\\ 
Slightly-minimal & yes & 0.5 & 0.001\\
Maximal & no & 0.5 & 0
\end{tabular}
\end{center}
\item   [Conservativity:] An agent can be {\em conservative} or {\em
non-conservative}.  They vary on how they calculate the probability
distribution over the expanded set of possible options when a
new variable gets added to the \dn.

All agents set $n_{v=i|j}$ to 0 when it discovers a new variable.
But the conservative and non-conservative agents vary on the how
they re-set the probability distribution over dependencies and the
Dirichlet $\alpha$-parameters:
\begin{center}
\begin{tabular}{@{}ll@{}}
& Estimating $Pr(\Pi^t_v)$ and $Pr(\theta_t)$\\
& when adding a new variable\\ \midrule
Conservative & Use (\ref{eq:old-parent-sets-new-variable}),
(\ref{eq:parenthood-new-variable}),
and (\ref{eq:alpha}) with $K=20$\\
Non-Con & Use (\ref{eq:prior-parenthoods}) and
$\alpha_{n=i|j}=\alpha_0=0.5$ 
\end{tabular}
\end{center}
\item [Domain Strategy:] The agent's domain-level strategy is to
execute an optimal action vs.\ some other action in a ratio of
$(1-\epsilon):\epsilon$, and initially $\epsilon=0.3$ (see
Section~\ref{sec:domain}).  We vary how this strategy changes during
learning as follows: (i) $\epsilon$ does not change (Static); or (ii)
$\epsilon$ gets progressively reduced (Decay).
\begin{center}
\begin{tabular}{@{}lll@{}}
& Initial $\epsilon$ & Decay factor \\
& & (per time step) \\ \midrule
Static & 0.3 & 1.0 \\
Decay & 0.3 & 0.999 
\end{tabular}
\end{center}
\end{description}
Unless stated otherwise, the agents we evaluate are
{\em Min-Vocab}, {\em Minimal}, {\em Conservative} and {\em Static}, and 
the expert has {\em Default Tolerance}.   

The {\bf baseline} agent for our task adopts a standard
$\epsilon$-greedy strategy (with $\epsilon$ set to 0.3) to learn an
optimal policy from $dn_0$: it ignores observed evidence that
makes $dn_0$ invalid.  So whatever the observed evidence, the
baseline agent's random variables ${\cal C}_0\cup {\cal A}_0$ and
dependencies $\Pi_0$ do not change.  All our other agents learn valid
\dn{}s (see Fact~\ref{fact:valid}).  We compare the performance of the
following nine agents on our task: the Baseline agent, Default (i.e.,
min-vocab, minimal, conservative, static, default tolerance),
Low-Tolerance (expert has low tolerance); Slightly-Min (agent is
slightly minimal), Maximal (agent is maximal), Non-Con (agent is
non-conservative), Decay (agent explores progressively less during
learning), and Full-Vocab (the agent starts out aware of $dn_+$'s
vocabulary, but not its dependencies or reward function).

In some of the 100 simulations, an agent that lacks the computational
advantages of Minimality---a principle we argued would enhance
tractability---crashes before it is observed and learned from 3000
pieces of evidence.\footnote{We define a crash as an ``Out of Memory''
  error on a Java Virtual Machine with a 4GB Memory Allocation Pool,
  or a simulation taking longer than its maximum allocated time of 12
  hours.} In that case, the agent's \dn{} just before the crash
occurred is used to compute the performance metric for that
simulation.

\subsection{Random \dns{}: Results}
\begin{figure}
\centering
\begin{subfigure}{0.80\textwidth}
\includegraphics[width=\textwidth]{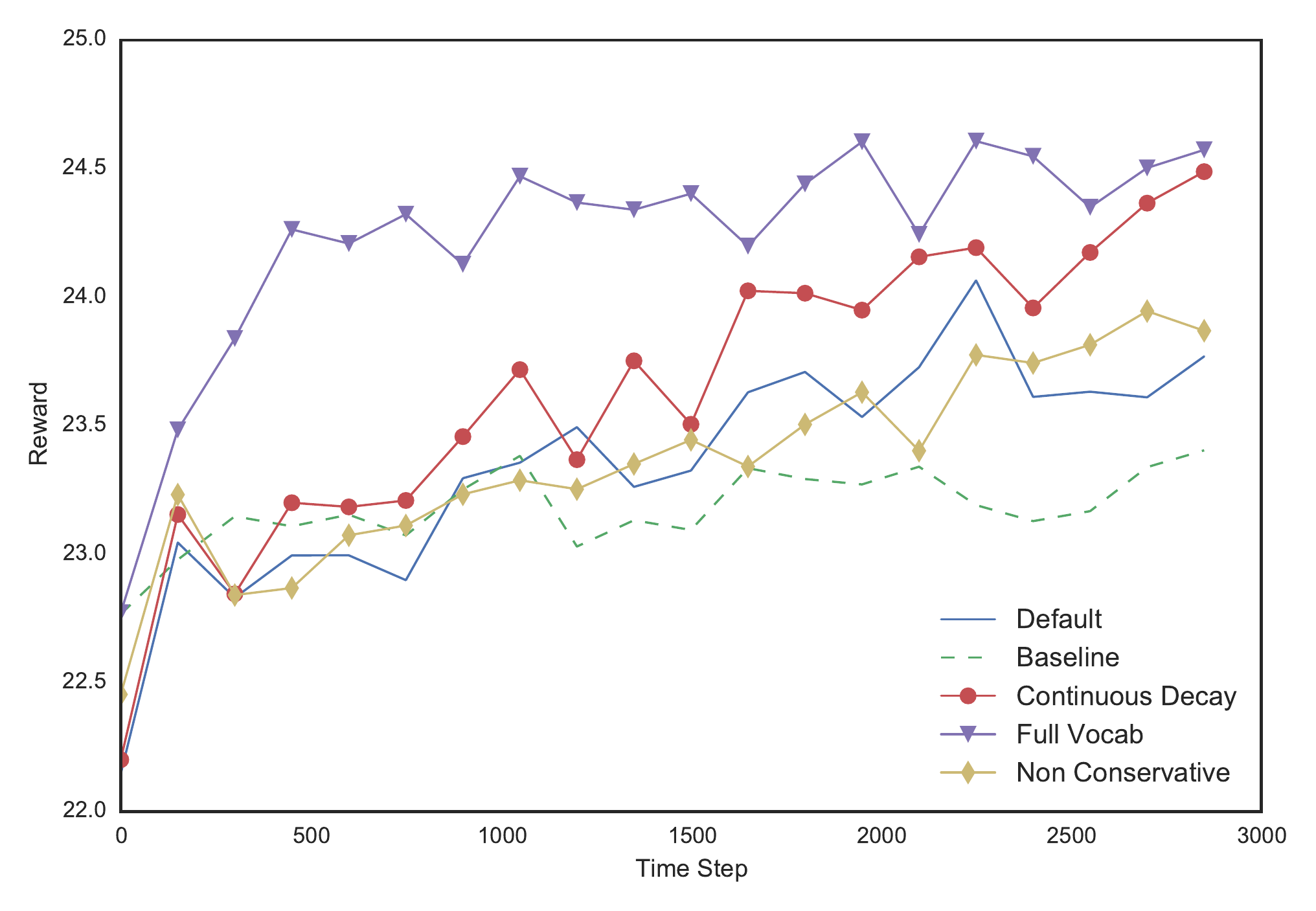}
\end{subfigure}

\begin{subfigure}{0.80\textwidth}
\includegraphics[width=\textwidth]{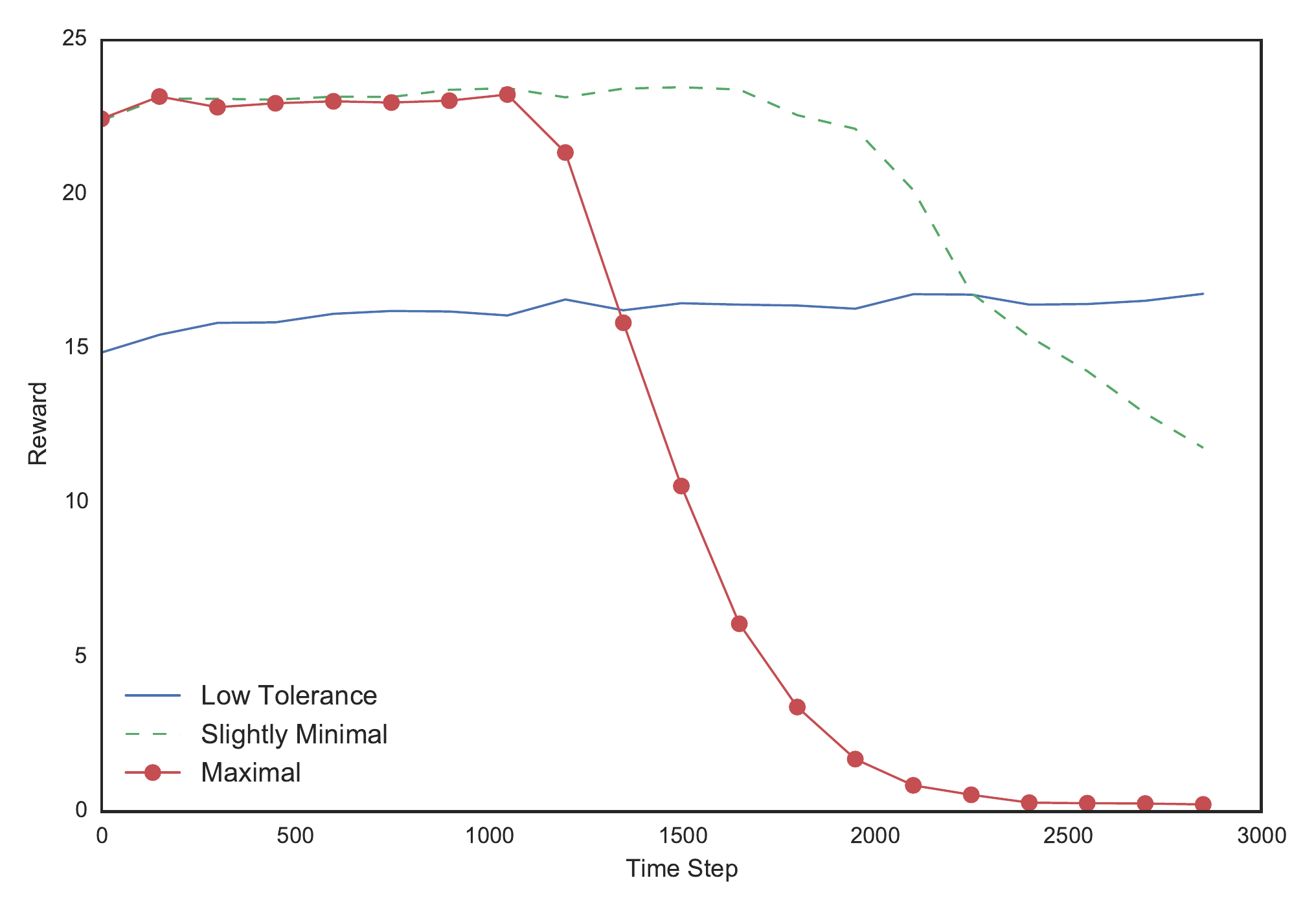}
\end{subfigure}
\caption{Rewards on $dn^{\mathit{best}}_+$. Results averaged across 100 simulations, and blocks of 150 time steps.  Note the scales in the two graphs differ, to help depict the performance of all agent types---the dips in the second figure are due to the agents crashing before t=3000.}
\label{fig:rewards_good}
\end{figure}

Across a variety of randomly generated \dns, each with different
dependencies, \cpt{}s, and reward domains, our agent converges to a
near-optimal policy with a relatively small amount of data (3000
pieces of evidence). Tables~\ref{table:policy_table} and
~\ref{table:cum_reward_table} show the performance metrics for each
agent in terms of final policy error (PE) and accumulated reward,
respectively.  Figure~\ref{fig:rewards_good} also shows the 
rewards gained from the domain trials {\em during} learning (averaged
across blocks of 150 domain trials and 100 simulations).
We report numbers for the \dn{} $dn^{\mathit{best}}_+$
where the agent was most successful at achieving a low PE, and for
$dn^{\mathit{worst}}_+$ where the agent was least successful, so as to
give a measure of the lower and upper bounds of the agent's
performance. (Full details on $dn^{\mathit{best}}_+$ and
$dn^{\mathit{worst}}_+$ are in the Appendix.) The p-values compare
each agent variation to the Default agent, and measure the probability
of seeing a result of this magnitude (or greater) given the null
hypothesis that the true average of both learning agents is identical.

From these results we see that the Default agent significantly
outperforms the Baseline agent, both in terms of final PE and total
accumulated reward.  Its PE is also significantly lower than the
Maximal and Non-Conservative agents.

\begin{table*}[t]
\centering
\begin{subtable}[t]{0.495\textwidth}
\centering
\ra{1.1}
\begin{tabular}{@{}llll@{}}
\toprule
& $PE_{t}$ & p-value \\ \midrule
Default & $0.34 $ & - \\
Baseline & $1.85$ & 0.00 \\
Low-Tol & $0.51 $ & 0.01 \\
Slightly-Min & $0.49$ & 0.01 \\
Decay & $0.47 $  & 0.03 \\
Full-Vocab & $0.08 $ & 0.00 \\
Non-Con & $0.78 $ & 0.00 \\
Maximal & $1.21$ & 0.00 \\ \bottomrule
\end{tabular}
\caption{$dn^{\mathit{best}}_+$}
\end{subtable}
\begin{subtable}[t]{0.495\textwidth}
\centering
\ra{1.1}
\begin{tabular}{@{}llll@{}}
\toprule
& $PE_{t}$ & p-value \\ \midrule
Default & $0.97$ & - \\
Baseline & $3.05$ & 0.00 \\ 
Low-Tol & $1.10$ & 0.20 \\
Slightly Min & $1.08$ & 0.24 \\
Decay & $0.86$ & 0.20 \\
Full-Vocab & $0.03$ & 0.00 \\
Non-Con & $1.44$ & 0.00 \\
Maximal & $1.30$ & 0.00 \\ \bottomrule
\end{tabular}
\caption{$dn^{\mathit{worst}}_+$}
\end{subtable}
\caption{Policy error (PE) for agents at t=3000. Average of 100 runs reported.} 
\label{table:policy_table}
\end{table*}

\begin{table*}[t]
\begin{subtable}{0.495\textwidth}
\centering
\ra{1.1}
\begin{tabular}{@{}llll@{}}
\toprule
& Reward & p-value \\ \midrule
Default & $70043$ & - \\
Baseline & $69541$ & 0.01 \\
Low-Tol & $48726$ & 0.00 \\
Slightly-Min & $62145$ & 0.00 \\
Decay & $70940$  & 0.00 \\
Full-Vocab & $72701$ & 0.00 \\
Non-Con & $70079$  & 0.85 \\
Maximal & $36760$  & 0.00 \\ \bottomrule
\end{tabular}
\caption{$dn^{\mathit{best}}_+$}
\end{subtable}
\begin{subtable}{0.495\textwidth}
\centering
\ra{1.1}
\begin{tabular}{@{}llll@{}}
\toprule
& Reward & p-value \\ \midrule
Default & $79894$ & - \\
Baseline & $79211$ & 0.00 \\
Low-Tol & $53568$ & 0.00 \\
Slightly-Min & $79711$ & 0.32 \\
Decay & $81518$ & 0.00 \\
Full-Vocab & $84342$ & 0.00 \\
Non-Con & $79397$ & 0.00 \\
Maximal & $77158$ & 0.00 \\ \bottomrule
\end{tabular}
\caption{$dn^{\mathit{worst}}_+$}
\end{subtable}
\caption{Accumulated reward for agents at t=3000. Average of 100 runs reported} 
\label{table:cum_reward_table}
\end{table*}

We now discuss the various factors that affect learning.

\paragraph{Initial awareness}
As expected, starting with complete knowledge of the vocabulary is an
advantage.  Tables~\ref{table:policy_table}
and~\ref{table:cum_reward_table} show that the Full-Vocab agent
achieves the lowest PE and highest cumulative rewards across all
simulations. This is unsurprising: the agent never has to ``throw away''
counts from the trials, and it deliberates over all actions from the
start.  However, the Full-Vocab agent takes significantly longer to
run than the Default agent, often anywhere from 3 to 8 times as long
(see Table~\ref{table:run_time_table}). There are two reasons for
this. First, inference over a \dn{} with more variables is usually
more expensive---the Full-Vocab agent infers a 21 variable \dn{} from
the start, while the Default agent initially updates a \dn{} with a
much smaller vocabulary.  Second, starting with a large initial
vocabulary and a fairly weak prior makes it more difficult for
Structural Update to prune the search space of parent sets. There is
often an initial explosion in the number of parent sets that the
Full-Vocab agent considers reasonable, because it has not yet gathered
enough information from the domain trials to discriminate unreasonable
parent sets from reasonable ones. In contrast, when the agent starts
out aware of fewer variables, each new variable is incrementally
introduced into a context where the agent has observed enough evidence
to construct a more aggressive prior via equation
(\ref{eq:old-parent-sets-new-variable}).

\begin{table*}[ht]
\begin{subtable}{0.495\textwidth}
\centering
\ra{1.1}
\begin{tabular}{@{}llll@{}}
\toprule
& Time (s) & p-value \\ \midrule
Default & $3081$ & - \\
Baseline & $13$  & 0.00 \\
Low-Tol & $8196$  & 0.00 \\
Slightly-Min & $12264$  & 0.00 \\
Decay & $2923$ & 0.30 \\
Full-Vocab & $12628$ & 0.00 \\
Non-Con & $2174$ & 0.00 \\ \bottomrule
\end{tabular}
\caption{$dn^{\mathit{best}}_+$.}
\end{subtable}
\begin{subtable}{0.495\textwidth}
\centering
\ra{1.1}
\begin{tabular}{@{}llll@{}}
\toprule
& Time (s) & p-value \\ \midrule
Default & $1266$  & - \\
Baseline & $15$  & 0.00 \\
Low-Tol & $2162$ & 0.00 \\
Slightly-Min & $4979$  & 0.00 \\
Decay & $1198$ & 0.31 \\
Full-Vocab & $9465$ & 0.00 \\
Non-Con & $1134$ & 0.02 \\ \bottomrule
\end{tabular}
\caption{$dn^{\mathit{worst}}_+$}
\end{subtable}
\caption{Run time in seconds for agents at t=3000. Average over 100 runs reported. (``Maximal" agent omitted, as none of the agent's runs managed to finish)}
\label{table:run_time_table}
\end{table*}

\begin{figure}
\centering
\includegraphics[width=0.9\textwidth]{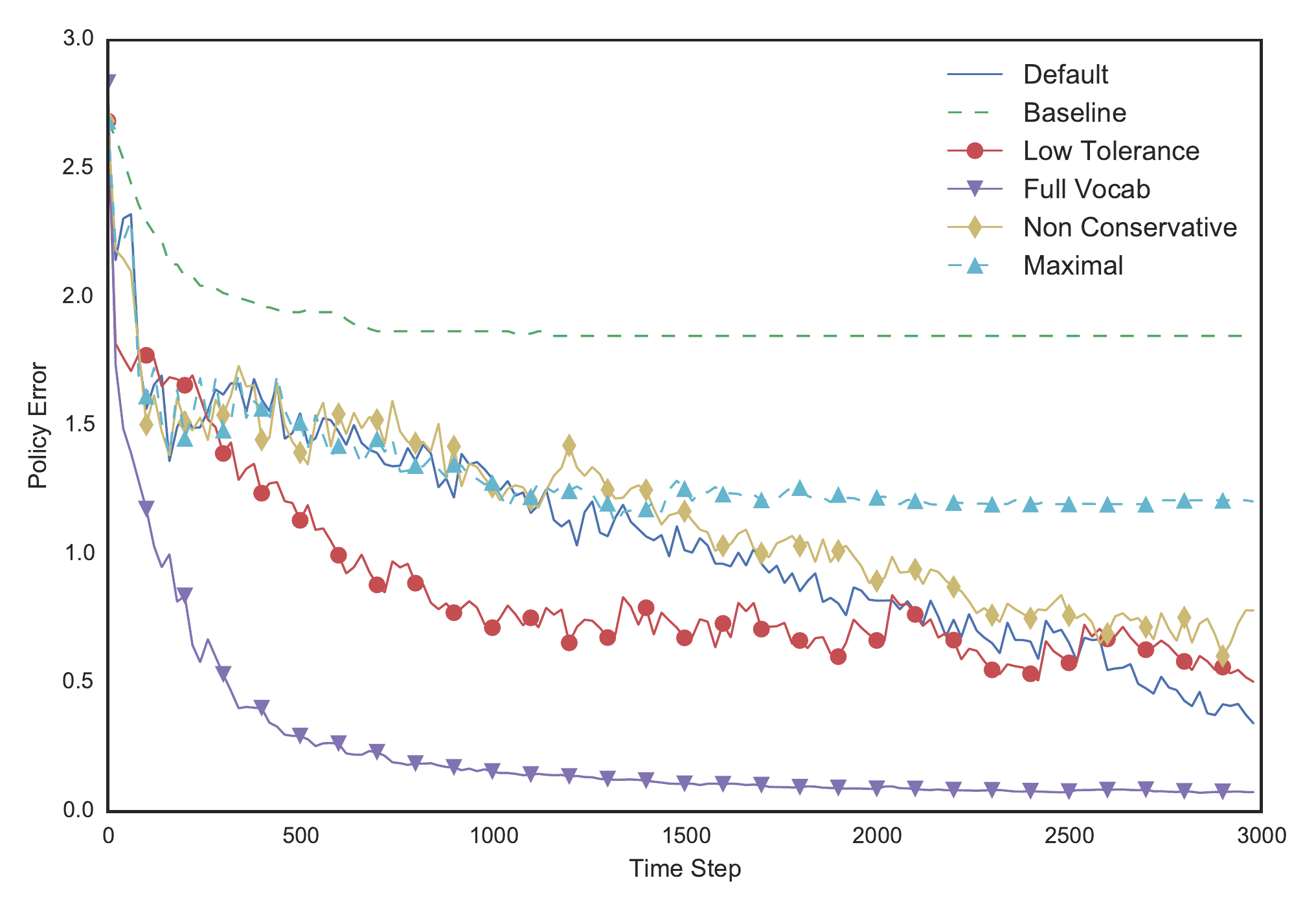}
\caption{Policy error at each time step on
$dn^{\mathit{best}}_+$ (averaged across
100 simulations).}
\label{fig:policy_error_graph}
\end{figure}

\paragraph{Expert tolerance}
In comparison to the Default expert, the Low-Tolerance expert results
in a marginally worse PE, and significantly worse total accumulated
reward. This is partly because each time the agent receives a message
from the expert, it misses both an opportunity to receive a reward by
taking an action in the domain, and also an opportunity to improve its
estimate of $\Pi$ and $\theta$ by observing an additional domain
trial. In our experiments, the Low Tolerance expert can end up sending
as many as 15 times more messages than the Default.  One advantage of
the Low-Tolerance expert, however, is that the agent makes early
improvements to its policy.  This is illustrated by Figure
\ref{fig:policy_error_graph}, which shows the PEs over time.

\paragraph{Minimality} The results show that at least some level of
minimality is necessary. In the vast majority of simulations (70\%--100\%, depending on $dn_+$), the Maximal agent runs out
of either time or memory, and so fails to collect rewards or improve
its policy via the 3000 pieces of evidence. By comparison, the Default
agent does not crash in any of the simulations.

Without a mechanism to prune the set of parent sets or the domain of the reward function, the agent must consider an exponential number of options.  For more than a small number of variables, this quickly becomes intractable.  The differences are less dramatic for the Slightly-Minimal agent: fewer simulations crash, and those that do tend to crash come later in the simulation. However, its PE is still significantly worse than the Default agent's PE. Excluding crashed agents from the results yields no significant difference in PE, suggesting that even with infinite time and resources, there is not necessarily any harm to adopting Minimality. The largest difference is in the run times: the Slightly Minimal Agent takes around 4 times as long to run as the Default agent. Thus, Minimality makes the problem tractable without harming performance.

\paragraph{Conservativity} The Non-Conservative agent has a
significantly worse PE than the Default agent. There appear to be two
benefits to preserving information about which parent sets are
unreasonable when a neologism is introduced.  First, it directs the
agent towards more accurate representations of the dependencies within
the \dn, and therefore hopefully allows it to make better decisions
with equivalent samples.  Second, having a more aggressive prior leads
the (Conservative) Default agent to learn more vocabulary---on average
19 versus 17---as it is more likely to encounter a context where it
asks the expert question (\ref{eq:which-effect}) (``What is affected
by X?''). This also explains why the Non-Conservative agent is
marginally faster than the Default agent---it reasons with a smaller
vocabulary.

\paragraph{Domain Strategy}
The Decay agent gathered a significantly larger total reward than the
Default agent, without a significant difference in their PEs. In other words, gradually reducing the extent to which the agent explores the
space as it gathers more evidence helps it to converge faster on a
decent policy.  But we haven't demonstrated here whether the Decay
agent would maintain this advantage as one increases the number
of variables of $dn_+$: it may be a disadvantage to gradually explore
less if the agent is more radically
unaware at the start of the learning process than we used in our
experiments here.  Additionally, while it was not within the
scope of these experiments to explore the space of exploration versus
exploitation strategies (we arbitrarily chose an initial split of
0.7:0.3), it is possible that superior initial parameters (and decay
rates) exist for learning the optimal policy.

\subsection{Barley \dn{}: Setup}

\begin{figure}
	\centering
	\begin{subfigure}[b]{\textwidth}
		\centering
		\includegraphics[scale=0.7]{barley.pdf}
		\caption{True \dn}
	\end{subfigure}
	\par\bigskip
	\begin{subfigure}[b]{\textwidth}
		\centering
		\includegraphics[scale=0.7]{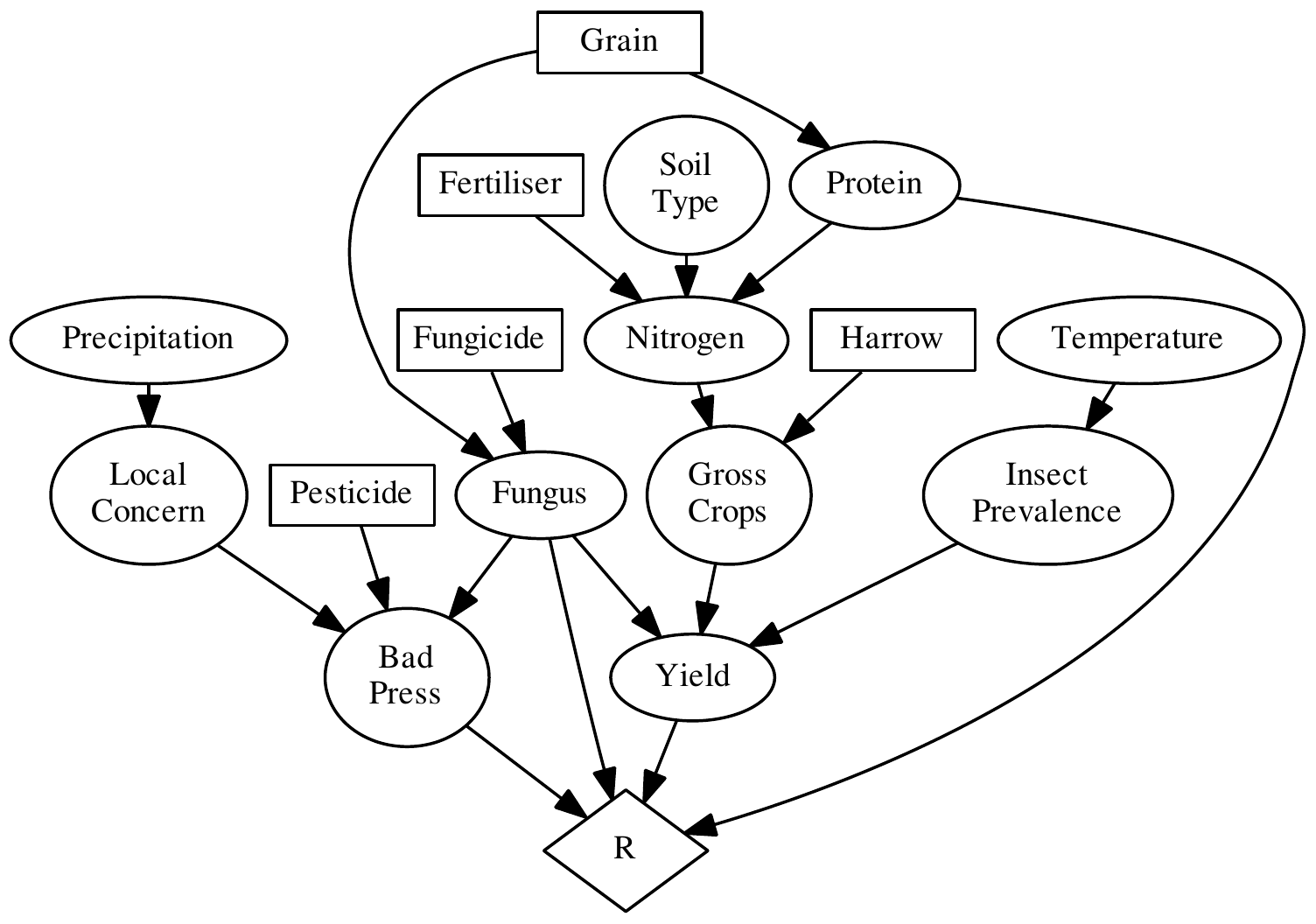}
		\caption{Learned \dn}
	\end{subfigure}
	\caption{Difference in structure between the true \dn{} and one learned by default agent at t=3000 for a (randomly chosen)  simulation}
	\label{fig:final_graphs}
\end{figure}

In these experiments, we test our \textbf{default} agent's performance
on the hand-crafted barley example (see Figure~\ref{fig:barley}) and
compare it against
the \textbf{baseline} agent (which ignores evidence that makes $dn_0$
invalid). The Barley \dn{} is intended to reflect a somewhat
realistic domain for our novel task, and has been explicitly designed
so that learning the optimal policy requires knowledge of factors that
the agent is initially unaware of.

As in the previous experiment, we run 100 simulations over 3000 pieces
of evidence then average the results. The structure of the agent's
initial decision network $dn_0$ and the true decision network $dn_+$
are shown in Figure~\ref{fig:barley}. The \cpt{}s for each node are in
the Appendix.

\subsection{Barley \dn{}: Results}

\begin{figure}
\centering
\begin{subfigure}[b]{0.49\textwidth}
\centering
\includegraphics[width=\textwidth]{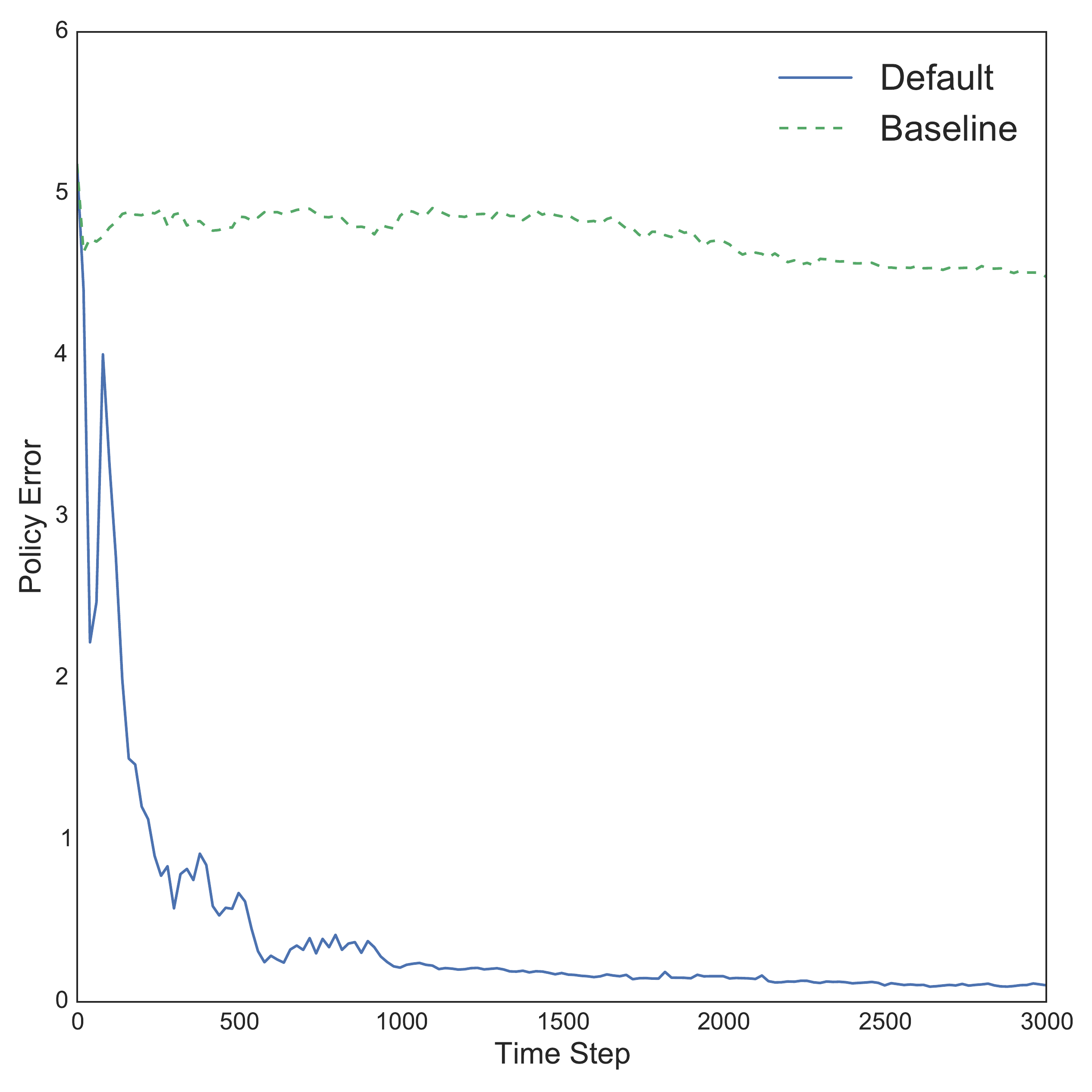}
\end{subfigure}
\begin{subfigure}[b]{0.49\textwidth}
\centering
\includegraphics[width=\textwidth]{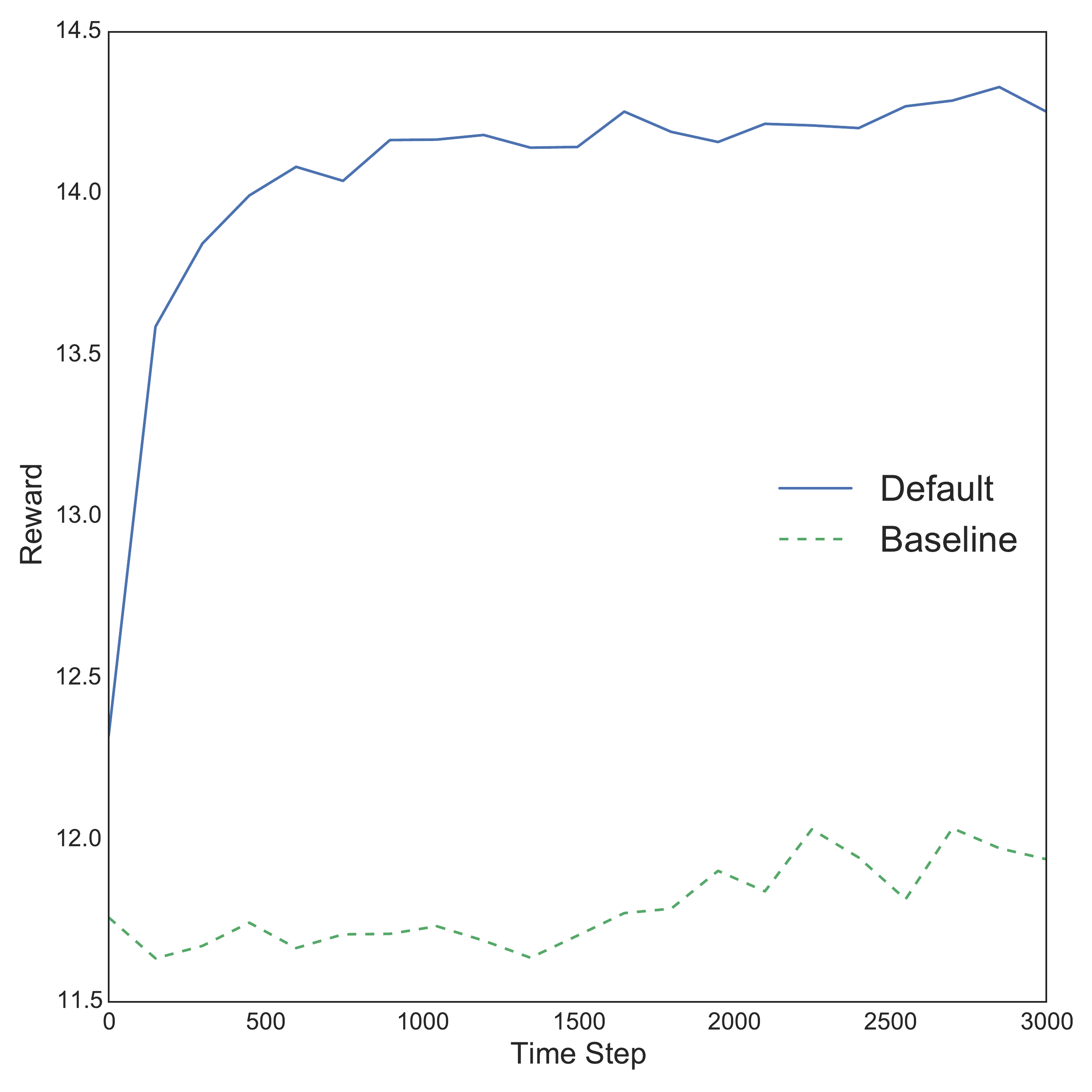}
\end{subfigure}
\caption{Reward and policy error for agents on the Barley task}
\label{fig:barleyResult}
\end{figure}

Figure \ref{fig:barleyResult} shows the policy error and average
reward over 3000 pieces of evidence for the Barley \dn{}. As expected,
the baseline quickly stops improving as it reaches the bounds of what
is achievable given its limited level of awareness. On the other hand,
the default agent continues to improve its policy as it discovers
unforeseen concepts and actions, and eventually converges upon a
near-optimal policy.

In addition to learning a near-optimal policy, the agent also learns a
fairly accurate and efficient representation of the underlying
decision network for the task. Figure \ref{fig:final_graphs} shows an
example of the agent's learned graph at t=3000. Notice that, in this
run, the agent did not manage to discover two pieces of the true
vocabulary---namely $Weeds$ and $\mbox{{\em Infestation}}$. Both of these are outcome variables which are not in the reward domain. In future work, we aim to explore methods to allow the agent discover these variables which provide a more efficient factorization of the problem if discovered, but which are not necessarily vital for learning the optimal policy.

Despite missing these two variables, the agent attempts to make
reasonable compensations for their absence. For example, in the
absence of the $\mbox{{\em Infestation}}$ variable, the agent assumes
that $\mbox{{\em Insect-Prevalence}}$ is directly
connected to $Yield$. The few other incorrect edges are either due to
reversals of causality (for instance, the model assumes $\mbox{{\em Protein}}$
influences $\mbox{{\em Nitrogen}}$ instead of vice versa) or by the model's bias
towards simplicity in the absence of sufficient domain trials (for
example, $\mbox{{\em Precipitation}}$ has a relatively weak influence on $Nitrogen$
compared to its many other parents, so the agent instead spuriously
connects $\mbox{{\em Precipitation}}$ to $\mbox{{\em Local-Concern}}$---a variable with no
parents).

\section{Related Work}
\label{sec:related-work}

 

The literature on learning optimal policies includes methods
for dealing with numerous types of change, such as non-stationary
rewards \nncite{Besbes:2014}, or noise in sensing and actuation
\nncite{daSilva:2006}.  Such methods have also been applied in
situations involving hidden variables (e.g., with Partially Observed
Markov Decision Processes \nncite{leonetti:2011}), or in cases where
the learner must reason about the underlying probabilistic structure of
the problem (e.g, with Factored Markov Decision Processes
\nncite{degris2006learning}). However, these methods require all
possible states and actions to be known in advance of learning. That
is, the agent starts out with complete knowledge of ${\cal C}$ and
${\cal A}$. In contrast, our agent learns these components from
evidence.

There are a few notable exceptions in the reinforcement learning (\rl)
literature, which attempt to relax the standard assumption that all
possible states and actions are known in advance of learning.
\namecite{rong:2016} defines Markov Decision Processes with
Unawareness ({\sc mdpu}s), and also provides algorithms for learning
optimal policies, even when the agent starts out unaware of actions
and states that influence what is optimal.  This work allows an agent
to discover an unforeseen atomic state or unforeseen action via a
random exploration function that is a part of the {\sc mdpu}.  Our
work contrasts with this in several ways.  First, we support
discovering unforeseen possibilities from evidence that stems from
communication with an expert, not just via exploration.  Secondly, and
in light of this quite different kind of evidence, the agent is not
simply discovering at most one new atomic state from the latest piece
of evidence, but rather discovering an unforeseen {\em concept}, which
instantly extends the set of atomic states considerably: e.g., an
unforeseen concept corresponding to a Boolean random variable doubles
the set of atomic states.  Accordingly, and in contrast to {\sc
  mdpu}s, the set of atomic states in our task is not defined by a
single random variable whose set of possible values change as the
agent becomes more aware.  Rather, the domain states are
conceptualised via a {\em set} of random variables in a dependency
structure, and the agent must learn that dependency structure so that
computing joint probability distributions in the domain remains
tractable.  Learning the concepts and their dependencies is
challenging but ultimately worth it, because the probabilistic
conditional independence they encapsulate makes learning and reasoning
(whether exact or approximate) more tractable.  This enables us to
conduct experiments in Section~\ref{sec:experiments} where the (true)
decision problem consists of 2 million distinct atomic states,
compared with the few thousand distinct atomic states in the
discretised models that are used in Rong's experiments.

\emph{Selective perception} methods, such as McCallum's
\nncite{mccallum1996reinforcement} U-tree algorithm,
were designed for problems in which the true state space may
technically be available, but is infeasibly large to reason with. The
U-tree method initially treats all states as indistinguishable, then
recursively splits states in two as the agent discovers significant
differences in their expected reward. The agent thus learns an
increasingly complex representation of the state space. The
U-tree algorithm handles sequential planning while our model
(currently) supports only single-stage decision problems; however, there are
three main learning tasks that our model supports that the U-tree
algorithm does not.  First, our model accommodates situations where
the agent is made aware of an entirely new possible action, while the
U-Tree algorithm assumes that the set of actions remains
fixed.  Secondly, our model learns from (qualitative) evidence provided by
a domain expert, as well as evidence from trial and error.  And
thirdly, our model learns unforeseen {\em concepts} and causal
relations among them, while the U-tree algorithm does not
reason about which dimensions define the state space, nor their causal
dependencies.  This third difference is related to the second: we
ultimately want our model to enable an agent to learn about the domain
and how to behave in it from a human expert, and it is quite natural for a
human to justify why one state is different from another on the basis
of the concepts on which optimal behaviour depends by saying, for
instance, ``sandy soil causes abundant barley yields''.  Our model
supports learning from this type of evidence.

Deep reinforcement learning (\drl{}, \namecite{mnih:etal:2015}) also
aims to learn a suitable abstraction of the data to enhance
convergence towards optimal policies. \drl{} combines deep learning
(i.e., a convolutional neural network or {\sc cnn}
\nncite{bengio:etal:2013,hinton:etal:2006}) with \rl{}, and has proven
extremely useful (e.g., \namecite{chouard:2016}). However, the major
successes in \drl{} have not so far considered changing state-action
spaces in the sense of the examples we have discussed. Instead, the
focus has been on using a large number (typically, in the millions or
tens of millions) of domain trials to re-describe a large state-action
space in terms of abstracted versions that render policy learning more
tractable. We believe that agents can learn more efficiently if they
exploit evidence from both domain trials and messages conveyed by an
expert. Unfortunately, expert evidence about causal relations (``X
causes Y'') or preference (``If C is true, then doing X is better than
doing Y'') are inherently symbolic. Adjusting weights in a
sub-symbolic representation like a {\sc cnn} so that it satisfies
certain qualitative properties is currently an unsolved
problem. Another limitation is that the implicit models generated by
\drl{} lack explainability---they are unable to elaborate on
\emph{why} a given action was recommended over another. Our approach
complements methods like \drl{} by dynamically building an
\emph{interpretable model} of the decision problem. Such interpretable
models make it easier for the agent to explain the reasons behind a
given decision, and also allow the agent to evaluate its
interpretation of expert messages against its current
conceptualisation of the decision problem via the standard logical
technique of model checking.

The agent must learn how to revise its model of the decision problem
to incorporate newly discovered actions and concepts. Hence, part of
our task is to learn the domain's causal structure.  Several
approaches exist for jointly learning causal dependencies and
probability distributions (e.g., \namecite{bramley:etal:2015,
  buntine:1991,friedman:1998}) and exploiting causal structure to
speed-up inference \nncite{ar2016jair}. We extend this work to meet
our objectives. First, models for learning dependencies all assume
that the vocabulary of random variables does not change during
learning, but in our task it changes when the agent becomes aware of
an unforeseen action or concept. Secondly, optimal action depends on
payoffs as well as beliefs, and so we must integrate learning
dependencies with learning potential payoffs.  Finally,
\namecite{bramley:etal:2015} and \namecite{friedman:1998} use only
domain trials as evidence, and although \namecite{buntine:1991} uses
expert evidence, the messages are only about causal relations and they
are all declared prior to learning.  In contrast, we want to {\em
  interleave} dialogue and learning: this enables the expert to offer
advice in a timely and contextually relevant manner; consequently, she
can explain particular outcomes and correct mistakes.

The \dn{}s used to model the decision problems in this paper resemble a simplified version of \emph{influence diagrams} \nncite{howard2005influence}. Influence diagrams allow one to express multi-stage decision problems by defining \emph{information arcs} between chance nodes and decision nodes, to help an agent assess whether it is worth observing a given trial or not (known as a \emph{value-of-information} calculation). There has been some work on inferring the probabilistic structure and utilities of other agents in influence diagrams \nncite{bielza2010modeling,nielsen2004learning,suryadi1999learning}. However, as explained in Section~\ref{sec:task}, the added complexity of the influence diagram definition was unnecessary to tackle the single-stage, fully observable tasks addressed in this paper. Moreover, in our work we assume that the agent knows its own utility function.

There are several areas of research in which expert evidence is used
to improve the performance of a learning agent. In \emph{Transfer
  learning}--- where knowledge of the optimal policy for a
\emph{source task} is used to improve performance on a related
\emph{target task}---the transfer of knowledge is sometimes captured
via an expert ``giving advice'' to a learner
\nncite{torrey2006relational}. Typically, both the source and target
task must belong to the same domain. If they do not, an explicit
mapping is usually provided from the states and actions in the source
task to the states and actions in the target task. In contrast, our
agent incrementally learns a single task, but is occasionally made
aware of new concepts, which it must learn to accommodate into its
existing knowledge. Another difference is that in transfer learning,
the expert advice is all declared before the agent begins learning. In
our model, the agent and expert engage in a dialogue throughout
learning.

\namecite{knox:stone:2009} use human expert evidence to inform \rl,
but they confine this evidence to updating the likely outcomes of
actions and their rewards; they do not support cases where expert
evidence reveals information that the agent {\em was not aware was
  possible}.  On the other hand, researchers have developed models for
learning optimal policies via a combination of domain actions and
natural language instructions, where the instructions may include
neologisms, whose semantics the agent must learn (e.g.,
\cite{liang:2005}).  But this work assumes that the neologism denotes
an already known concept within the agent's abstract planning
language. On encountering the neologism ``block'' , for example, the
agent's task is to learn that it maps to the concept {\tt block} that
is already an explicit part of the agent's domain model.  In contrast,
we support learning optimal policies when the neoglogism denotes a
concept that the agent is currently {\em unaware} of; e.g., to support
learning when {\tt block} is not a part of the agent's
conceptualisation of the domain, and yet this (unforeseen) concept is
critical to optimal behaviour.

\namecite{forbes:etal:2015} use embodied natural language instructions
to support teaching a robot a new skill (a task known as {\em learning
  by demonstration}).  Their model learns how to map a natural
language neologism to what might be a novel combination of sensory
values---in this sense, the meaning of the neologism may be an
unforeseen concept---and this novel concept then informs the task of
learning new motor controls.  In effect, this work links a rapidly
growing body of research on learning how to ground natural language
neologisms in the embodied environment (known as the symbol grounding
problem;
\nncite{siskind:1996,dobnik:etal:2012,yu:etal:2016,hristov:etal:2017}
{\em inter alia}) with learning a new skill.  But it does not support
{\em integrating} the newly acquired skill into an existing hypothesis
space consisting of other actions, consequences and rewards; so they
do not support learning optimal policies when not only this new skill
but other actions (and related concepts) are needed as well.

Our aim in this paper is to supply a complementary set of learning
algorithms to this prior work.  Like \cite{forbes:etal:2015}, our
agent learns from both trial and error and from instruction.  But we
focus on a complementary task: instead of focussing, as they do, on
{\em learning how} to execute a new skill, we focus on {\em learning
  when} it is optimal to execute it.  Given that we wish to support
larger planning tasks, we also broaden the goals to discovering and
learning to exploit arbitrary unforeseen concepts, not just the
learning of spatial concepts that \namecite{forbes:etal:2015} are
limited to (see also \cite{dobnik:etal:2012}).  On the other hand,
natural language ambiguity makes extracting the hidden message from
natural language utterances a highly complex process
\nncite{grice:1975,bos:etal:2004,zettlemoyer:collins:2007,reddy:etal:2016}.
We bypass this complexity by using a {\em formal} language as the
medium of conversation.  This formal language can be broadly construed
as the kind of language one uses to represent the output of natural
language semantic parsing \nncite{artzi:zettlemoyer:2013}.  Replacing
the expert that we deploy in our experiments with an expert {\em
  human} would require linking the model we present in this paper with
existing work on grounded natural language acquisition and
understanding. This task is beyond the scope of the current paper, and
forms a major focus of our future work.

\namecite{lakkaraju2017identifying} tackle learning ``unknown
unknowns'' via interaction with an oracle. This work addresses a
specific type of unawareness: an agent assigns an incorrect label to a
trial with high confidence. Crucially however, they assume that the
correct label for this trial must be a label that the agent is already
currently aware of.  In other words, they exclude the option that the
hypothesis space itself is incomplete, which is the type of
unawareness that we are interested in. 

There is a growing body of work on modelling unaware agents (e.g.,
\namecite{feinberg:2004,halpern:rego:2013,board:etal:2011,heifetz:etal:2013}).
These theories predict when the agent's unawareness causes it to
deviate from what is actually optimal.  But with the exception of
\namecite{rong:2016}, this work does not address how an agent {\em
  learns from evidence}: i.e., how to use evidence to become aware of
an unforeseen factor and to estimate how this gets incorporated into
its updated conceptualisation of the decision problem.  We fill this
gap.  This prior work interprets awareness with respect to models that
include every possible option.  A fully aware agent (e.g., the domain
expert in our experiments) or an analyst can model an unaware agent
this way, but it does not characterise the unaware agent's {\em own
  subjective perspective} \nncite{li:2008}.  In contrast, our learning
task requires the unaware agent to use evidence to change its set of
possible options: it dynamically constructs this set and its causal
structure from evidence.

Finally, the barley example from Section~\ref{sec:intro}
shows that becoming aware of an unforeseen possibility can prompt
revisions, rather than refinements, to the reward function.  But the
models that support preference revision on discovering an unforeseen
possibility assume a qualitative preference model
\nncite{hansson:1995,cadilhac:etal:2015}.  Following
\namecite{cadilhac:etal:2015}, we use evidence to dynamically
construct an ever more specific partial description of preferences,
and defeasible reasoning yields a complete preference model from the
(partial) description---the agent defaults to indifference.  But
unlike \namecite{cadilhac:etal:2015}, the learning agent uses evidence
to estimate numeric payoffs rather than qualitative preference
relations.

\section{Conclusions and Future Work}
\label{sec:conclusion}

This paper presents a method for discovering and learning to exploit unforeseen possibilities while attempting to converge on optimal behaviour in single-stage decision problems.  The method supports learning by trial and error as well as learning by instruction via messages from a domain expert. We model the problem as one of dynamically building an {\em interpretable conceptualisation} of the domain: we argued that building an interpretable model offers a straightforward way to exploit both the {\em qualitative} evidence and the {\em quantitative} evidence.  Specifically, our model is the first to support learning {\em all} components of a decision network, including its set of random variables.

Our algorithms build upon the common sense notions of Minimality and
Conservativity.  Minimality is encoded in the
following ways: greedy search for a minimal vocabulary and domain for
the reward function; defaulting to indifference; defeasibly low prior
probabilities of dependencies; and the assumption that any superset of
a sufficiently improbable parent set is at least as improbable.
Conservativity is encoded by retaining the relative likelihoods among
dependencies and \cpt{}s
learned so far when a new variable gets added to the \dn.

Our experiments show that these principles significantly enhance the performance of the agent in its learning task, as measured by policy error and cumulative reward, when learning a \dn{} of 21 random variables from 3000 pieces of evidence. The model was also shown to be robust to the topology of the true \dn{}: in other words, it learns effective policies, whatever the true conceptualisation of the domain might be. An agent that starts with a minimal subset of the true hypothesis space is also competitive with one that starts with full knowledge of the set of random variables that define the atomic states.

There are several novel formal components to our model.  For instance,
we have combined symbolic constraint solving with Dirichlet distribution based
estimates to guide learning, and the novel ILP step ensures that the
estimated causal structure of the domain is {\em valid}---i.e., all
nodes are connected to the utility node, and the dependencies satisfy
all observed evidence to date.  This component on its own presents an
important extension to existing work on learning dependencies, since
it supports learning a \dn{} as opposed to just a Bayesian network.

These are just the first steps towards building an agent that can
learn to conceptualise and exploit its domain via its own action and via messages
from teachers or experts.  We would like to
extend this model to handle decision
problems where some of the critical factors for solving the task have
hidden values even {\em after} the agent becomes aware of them; this
would involve replacing our Parameter Update algorithm with a modified
version of Structural EM \nncite{friedman:1998}.  We
would also like to extend it to handle dynamic environments: i.e., to
learn (factored) {\sc mdp}s, where the \dn{}s learned here correspond
to just one time slice.  This would not only extend the scope of this
work from single-stage decision making to sequential planning; it
would also be essential if we
were to replace {\em defining} the agent's and expert's dialogue
strategy with {\em learning} an optimal dialogue strategy.  We would
also like to enrich dialogue in three ways.  First, we would like the learner
to utilise defeasible messages about likelihood.  Second, we would
like to relax the assumption that the expert's intended message is
always true.  And finally, we would like to support dialogues in
natural language---the latter two enhancements are a necessary step
for an agent to discover and exploit
unforeseen possible options via interaction with a human teacher.

\bibliography{revisedPaper}

\newpage
\section*{Appendix}

The model for learning $dn_+$ made use of a language ${\cal L}$ for
partially describing \dn{}s.  Its syntax and semantics is defined as
follows:

\begin{defn}{{\bf The syntax of the language ${\cal L}$}}\\
\label{defn:syntax of the description language}
\begin{itemize}
\item Terms of various sorts: $X,Y\ldots$ are {\em random variable}
(RV) constants; $\Pi_Y$, ${\cal V}$, ${\cal A}$, ${\cal O}$, ${\cal
B}$ are Sets of Random Variables (SRV) constants (denoting sets of
random variables in the model); where $X$ is an RV constant, $x$ is
a proposition term (we also say that $\neg x$ is a propositional
term); $s$ is an atomic state (AS) term (denoting an 
atomic state in the model).
The language also includes RV variables and AS variables.
\item   If $p$ is a propositional term or an AS term, then it is a well-formed
formula ({\sc wff}) in
${\cal L}$.
\item   If $p$ is a conjunction of positive and negative propositional
terms and
${\cal X}$ is an SRV constant, then $p\upharpoonright {\cal X}$ is a
well-formed formula ({\sc wff}) in ${\cal L}$. 
\item If $X$ is an RV term and ${\cal Y}$ is an SRV term, then
$X\in {\cal Y}$ is a {\sc wff} in ${\cal L}$.
\item   If $s$ is an AS term and $n$ is a number, then ${\cal R}(s)=n$
is a {\sc wff} in ${\cal L}$. 
\item   Where $\phi,\psi$ are {\sc wff}s, so are $\neg \phi$, $\phi
\wedge \psi$, $\exists s\phi$, $\forall s\phi$ (where $s$ is an AS variable
variable) and $?\lambda V\phi$ (where $V$ is an RV variable). 
\end{itemize}
\end{defn}

Each model $dn$ for interpreting ${\cal L}$ corresponds to a
(unique) complete \dn{} (see Definition~\ref{defn:dn}).  
Definition~\ref{defn:semantics}
then evaluates the formulae of ${\cal L}$ as partial descriptions of
$dn$.

\begin{defn}{{\bf The semantics of ${\cal L}$}}\\
\label{defn:semantics}
Let $dn=\langle {\cal C}_{dn},{\cal A}_{dn},\Pi_{dn},\theta_{dn},{\cal
R}_{dn}\rangle$ be a \dn{} and $g$ a variable assignment function.
\begin{itemize}
\item   For an RV constant $X$, $\sem{X}^{\langle
dn,g\rangle}=X$;
similarly for SRV constants.\footnote{If $X\not\in {\cal
C}_{dn}\cup {\cal A}_{dn}$, then $\sem{X}^{\langle dn,g\rangle}$
is undefined and Definition~\ref{defn:semantics}
ensures that 
any formula $\phi$ featuring $X$ is such that
$dn\not\models \phi$; similarly for
propositional terms $p$ featuring a value of a variable that is
not a part of $dn$.}  This ensures that $\sem{{\cal A}}^{dn,g}$,
$\sem{{\cal B}}^{dn,g}$ and $\sem{{\cal O}}^{dn,g}$ denote sets of
variables in the appropriate position in the causal structure of
$dn$. 
\item   For an AS variable $s$, $\sem{s}^{\langle
dn,g\rangle}=g(s)$ where $g(s)\in 2^{{\cal C}_{dn}\times {\cal
A}_{dn}}$. For an RV variable $V$, $\sem{V}^{\langle
dn,g\rangle}=g(V)\in {\cal C}_{dn}\cup {\cal
A}_{dn}$. 
\item   For an RV term $a$ and an SRV term $b$, $\sem{a\in
b}^{\langle dn,g\rangle}=1$ 
iff $\sem{a}^{\langle dn,g\rangle}\in \sem{b}^{\langle
dn,g\rangle}$.  
\item Where $p$ is a propositional term, $\sem{p}^{\langle
dn,g\rangle}=p$.  
\item  For an AS term $s$ and number $n$, $\sem{{\cal
R}(s)=n}^{\langle dn,g\rangle}=1$ iff ${\cal 
R}_{dn}(\sem{s})=n$. 
\item Where $p$ is a conjunction of positive and negative
propositional terms and 
${\cal X}$ is a set 
constant, $\sem{p\upharpoonright {\cal X}}^{\langle
dn,g\rangle}=\sem{p}^{\langle dn,g\rangle}\upharpoonright \sem{{\cal
X}}^{\langle dn,g\rangle}$ (i.e., the projection of the
denotation of $p$ onto the set of variables denoted by
${\cal X}$). 
\item   For formulae $\phi$, $\psi$: $\sem{\phi\wedge\psi}^{\langle dn,g\rangle} = 1$ iff
$\sem{\phi}^{\langle dn,g\rangle}=1$ and $\sem{\phi}^{\langle dn,g\rangle}=1$;
$\sem{\neg \phi}^{\langle dn,g\rangle}=1$ iff $\sem{\phi}^{\langle
dn,g\rangle}=0$;
$\sem{\exists s\phi}^{\langle dn,g\rangle}=1$ iff there is a variable
assignment function $g'=g[s/p]$ such that
$\sem{\phi}^{\langle dn,g'\rangle}=1$. 
\item Where $V$ is a RV variable and $\phi$ is a formula:\\
$\sem{?\lambda V\phi}^{\langle dn,g\rangle}=\{\phi[V/X]: X \mbox{ is
an RV constant and } \sem{\phi[V/X]}^{\langle dn,g\rangle}=1\}$.
\end{itemize}
\end{defn}
These interpretations yield a satisfaction relation in the usual way:
$dn\models \phi$ iff there is a function $g$ such that 
$\sem{\phi}^{\langle dn,g\rangle}=1$.

\subsection*{Derivation of Parameter Update Given a Domain Trial}

We now show that
(\ref{eq:corrected-parameter-update})---the formula for incrementally
updating the probability of 
a parent set given the latest domain trial---follows from the
recursive $\Gamma$ function in the Dirichlet distribution.  
\begin{equation*}
\begin{array}{l}
Pr(\Pi_v | e_{0:t}) 
= Pr(\Pi_v | e_{0:t-1})\frac{(n_{v=i|j} +
\alpha_{v=i|j} - 1)}{(n_{v=.|j} + \alpha_{v-.|j} - 1)}
\end{array}
\tag{\ref{eq:corrected-parameter-update}}
\end{equation*}
Exploiting the Dirichlet
distribution to compute the posterior probability yields the following, where
$m_v$ is the number of possible values for variable $V$ (so in our
case, $m_v=2$):
\[
Pr(\Pi_v|e_{0:t})\propto  Pr(\Pi_v)\prod_{j\in
\mathit{values}(\Pi_v)}\frac{Beta_{m_v}(n_{v=1|j}+\alpha_{v=1|j},\ldots n_{v=m_v|j}+\alpha_{v=m_v|j})}
{Beta_{m_v}(\alpha_{v=1|j},\ldots, \alpha_{v=m_v|j})}
\]
If $e_t$ makes $V=i$, $\Pi_v=j$, then the only term in the product
that differs between updating with $e_{0:t-1}$ vs.\ $e_{0:t}$ is when
$\Pi_v=j$.  For $e_{0:t-1}$, this term is:
\[
\frac{Beta_{m_v}(n_{v=1|j}+\alpha_{v=1|j},\ldots,n_{v=i|j}-1+\alpha_{v=i|j},\ldots,n_{v=m_v|j}+\alpha_{v=m_v|j})}{
Beta_{m_v}(\alpha_{v=1|j},\ldots,\alpha_{v=m_v|j})}
\]
So 
\[
\begin{array}{r l}
Pr(\Pi_v|e_{0:t})\propto  & Pr(\Pi_v|e_{0:t-1})\frac{Beta_{m_v}(n_{v=1|j}+\alpha_{v=1|j},\ldots n_{v=m_v|j}+\alpha_{v=m_v|j})}
{Beta_{m_v}(n_{v=1|j}+\alpha_{v=1|j},\ldots,n_{i|j}-1+\alpha_{v=i|j},\ldots,n_{m_v|j}+\alpha_{v=m_v|j})} 
\\
\multicolumn{2}{l}{= Pr(\Pi_v|e_{0:t-1})\frac{\prod_{k\in
\mathit{values}(V)}\Gamma(n_{v=k|j}+\alpha_{v=k|j})}{\Gamma(n_{v=.|j}+\alpha_{v=.|j})}
\frac{\Gamma(n_{v=.|j}-1+\alpha_{v=.|j})}{\Gamma(n_{v=i|j}-1+\alpha_{v=i|j})\prod_{k\in
\mathit{values}(V)\setminus i}\Gamma(n_{v=k|j}+\alpha_{v=k|j})}}\\
=  &
Pr(\Pi_v|e_{0:t-1})\frac{\Gamma(n_{n=i|j}+\alpha_{v=i|j})}{\Gamma(n_{v=.|j}+m_v\alpha_{v=.|j})}
\frac{\Gamma(n_{v=.|j}-1+\alpha_{v=.|j})}{\Gamma(n_{v=i|j}-1+\alpha_{v=i|j})}
\end{array}
\]
By the definition of $\Gamma$ (i.e., $\Gamma(\alpha) = (\alpha
-1)\Gamma(\alpha-1)$), we have:
\[
\begin{array}{r l}
Pr(\Pi_v|e_{0:t}) = &
Pr(\Pi_v|e_{0:t-1})\frac{(n_{v=i|j}-1+\alpha_{v=i|j})\Gamma(n_{v=i|j}-1+\alpha_{v=i|j})}{(n_{v=.|j}-1+\alpha_{v=.|j})\Gamma(n_{v=.|j}-1+\alpha_{v=.|j})} 
\frac{\Gamma(n_{v=.|j}-1+\alpha_{v=.|j})}{\Gamma(n_{v=i|j}-1+\alpha_{v=i|j})}\\
= &
Pr(\Pi_v|e_{0:t-1})\frac{(n_{v=i|j}-1+\alpha_{v=i|j})}{(n_{v=.|j}-1+\alpha_{v=.|j})}
\end{array}
\]
$\blacksquare$

\subsection*{Decision Networks Used in Experiments}

Tables \ref{tab:cpt-a}, \ref{tab:cpt-b}, and \ref{tab:cpt-c} specify the true conditional probability tables (\cpt{}s) and reward functions for the three Barley \dns{} used in the experiments of section \ref{sec:experiments} ($dn_+^{best}$, $dn_+^{worst}$, and $dn_+^{barley}$). For each \cpt, each row lists a variable's name, its immediate parents, and the probability of that variable having the value 1 given each possible assignment to the variable's parents (assignments are enumerated according to the order specified in the second column). The reward table lists the reward domain and the reward received given each possible assignment to the domain.

\begin{table*}[h]
\caption{$dn_+^{best}$}
\centering
\begin{subtable}{\textwidth}
\centering
\ra{1.4}
\begin{tabular}{@{}l|l|l@{}}
\toprule
$X$ & $\Pi_{X}$ & $P(X = 1\ | \Pi_{X})$ \\ \midrule
B1 & B5 & 0.596, 0.774 \\
B2 & $\emptyset$ & 0.653 \\
B3 & B7 & 0.457, 0.457 \\
B4 & $\emptyset$ & 0.354 \\
B5 & B3, B4 & 0.639, 0.224, 0.35, 0.273 \\
B6 & $\emptyset$ & 0.738 \\
B7 & $\emptyset$ & 0.313 \\
O1 & O2, O6, A3 & 0.286, 0.478, 0.401, 0.956, 0.53, 0.084, 0.766, 0.923 \\
O2 & A1, A7 & 0.31, 0.554, 0.213, 0.197 \\
O3 & A2, A5, B2 & 0.74, 0.945, 0.92, 0.721, 0.371, 0.963, 0.129, 0.029 \\
O4 & B2, O5, B1 & 0.937, 0.93, 0.484, 0.255, 0.637, 0.191, 0.136, 0.149 \\
O5 & O7 & 0.36, 0.43 \\
O6 & A6, B5 & 0.677, 0.209, 0.696, 0.521 \\
O7 & A4, B6 & 0.821, 0.379, 0.211, 0.383 \\ \bottomrule
\end{tabular}
\end{subtable}

\vspace{4mm}

\begin{subtable}{\textwidth}
\centering
\ra{1.4}
\begin{tabular}{@{}l|p{10cm}@{}}
\toprule
$\Pi_{R}$ & $\mathcal{R}(\Pi_{R})$ \\ \midrule
O1, O4, O3, O5, B7 & 43.55, 27.42, 35.48, 8.06, 20.97, 11.29, 30.65, 50.0, 46.77, 6.45, 12.9, 32.26, 41.94, 0.0, 33.87, 37.1, 25.81, 4.84, 22.58, 29.03, 14.52, 9.68, 45.16, 24.19, 17.74, 19.35, 48.39, 16.13, 40.32, 3.23, 38.71, 1.61 \\ \bottomrule
\end{tabular}
\end{subtable}
\label{tab:cpt-a}
\end{table*}

\begin{table*}[h]
\caption{$dn_+^{worst}$}
\centering
\begin{subtable}{\textwidth}
\centering
\ra{1.4}
\begin{tabular}{@{}l|l|l@{}}
\toprule
$X$ & $\Pi_{X}$ & $P(X = 1\ | \Pi_{X})$ \\ \midrule
B1 & $\emptyset$ & 0.49 \\
B2 & B3, B7 & 0.779, 0.547, 0.727, 0.197 \\
B3 & $\emptyset$ & 0.198 \\
B4 & $\emptyset$ & 0.36 \\
B5 & $\emptyset$ & 0.883 \\
B6 & $\emptyset$ & 0.237 \\
B7 & B1 & 0.43, 0.992 \\
O1 & O7, A1, A3 & 0.088, 0.467, 0.13, 0.548, 0.7, 0.372, 0.498, 0.047 \\
O2 & B5, O7 & 0.461, 0.599, 0.806, 0.37 \\
O3 & A6 & 0.881, 0.125 \\
O4 & O3, A5, A4 & 0.111, 0.033, 0.315, 0.322, 0.034, 0.579, 0.94, 0.644 \\
O5 & O2, B4, O4 & 0.816, 0.979, 0.577, 0.467, 0.459, 0.751, 0.191, 0.541 \\
O6 & O4, A2 & 0.818, 0.583, 0.188, 0.957 \\
O7 & O6, A7, B1 & 0.314, 0.418, 0.48, 0.822, 0.957, 0.889, 0.697, 0.061 \\ \bottomrule
\end{tabular}
\end{subtable}

\vspace{4mm}

\begin{subtable}{\textwidth}
\centering
\ra{1.4}
\begin{tabular}{@{}l|p{10cm}@{}}
\toprule
$\Pi_{R}$ & $\mathcal{R}(\Pi_{R})$ \\ \midrule
O1, B6, B7, O5, B2 & 43.55, 6.45, 22.58, 46.77, 37.1, 48.39, 1.61, 8.06, 16.13, 29.03, 50.0, 9.68, 11.29, 14.52, 19.35, 40.32, 4.84, 12.9, 38.71, 20.97, 45.16, 3.23, 32.26, 17.74, 33.87, 25.81, 24.19, 35.48, 27.42, 30.65, 41.94, 0.0 \\ \bottomrule
\end{tabular}
\end{subtable}
\label{tab:cpt-b}
\end{table*}

\begin{table*}[h]
\caption{$dn_+^{barley}$}
\centering
\begin{subtable}{\textwidth}
\centering
\ra{1.4}
\begin{tabular}{@{}l|p{6cm}|p{5cm}@{}}
\toprule
$X$ & $\Pi_{X}$ & $P(X = 1\ | \Pi_{X})$ \\ \midrule
Soil Type & $\emptyset$ & 0.5 \\
Temperature & $\emptyset$ & 0.5 \\
Precipitation & $\emptyset$ & 0.5 \\
Insect-Prevalence & $\emptyset$ & 0.5 \\
Local-Concern & $\emptyset$ & 0.5 \\
Nitrogen & Soil Type, Precipitation, Pesticide, Fertiliser & 0.4, 0.6, 0.5, 0.7, 0.3, 0.5, 0.4, 0.6, 0.65, 0.85, 0.75, 0.95, 0.55, 0.75, 0.65, 0.85 \\
Gross Crops & Harrow, Nitrogen, Grain & 0.5, 0.4, 0.8, 0.7, 0.6, 0.5, 0.9, 0.8 \\
Fungus & Temperature, Fungicide, Grain & 0.2, 0.5, 0.02, 0.04, 0.3, 0.6, 0.03, 0.06 \\
Weeds & Temperature, Harrow, Soil Type & 0.2, 0.1, 0.02, 0.01, 0.3, 0.15, 0.03, 0.015 \\
Infestation & Insect-Prevalence, Pesticide & 0.1, 0.5, 0.01, 0.05 \\
Yield & Gross Crops, Fungus, Weeds, Infestation & 0.2, 0.95, 0.1, 0.5, 0.1, 0.7, 0.05, 0.3, 0.05, 0.65, 0.01, 0.2, 0.01, 0.45, 0.005, 0.1 \\
Protein & Nitrogen, Fertiliser, Grain & 0.5, 0.9, 0.4, 0.8, 0.4, 0.8, 0.3, 0.7 \\
Bad Press & Local-Concern, Pesticide & 0.0, 0.0, 0.01, 0.5 \\

\bottomrule
\end{tabular}
\end{subtable}

\vspace{4mm}

\begin{subtable}{\textwidth}
\centering
\ra{1.4}
\begin{tabular}{@{}l|l@{}}
\toprule
$\Pi_{R}$ & $\mathcal{R}(\Pi_{R})$ \\ \midrule
Yield, Protein, Fungus, Bad Press & 10, 15, 15, 20, 0, 5, 5, 10, -10, -5, -5, 0, -20, -15, -15, -10 \\
\bottomrule
\end{tabular}
\end{subtable}
\label{tab:cpt-c}
\end{table*}


\end{document}